\documentclass{ecai} 

\usepackage{latexsym}
\usepackage{amssymb}
\usepackage{amsmath}
\usepackage{amsthm}
\usepackage{booktabs}
\usepackage{enumitem}
\usepackage{graphicx}
\usepackage{color}

\usepackage{multirow}
\usepackage{amsfonts}
\usepackage{subfig}
\usepackage{newfloat}
\usepackage{listings}

\usepackage[table,xcdraw]{xcolor} 
\usepackage{booktabs} 

\newtheorem{definition}{Definition}

\newcommand{\BibTeX}{B\kern-.05em{\sc i\kern-.025em b}\kern-.08em\TeX}

\begin{document}


\begin{frontmatter}


\paperid{5841} 


\title{Text-Attributed Graph Anomaly Detection via Multi-Scale Cross- and Uni-Modal Contrastive Learning}


\author[A,B]{\fnms{Yiming}~\snm{Xu}}
\author[A,B]{\fnms{Xu}~\snm{Hua}}
\author[A,B]{\fnms{Zhen}~\snm{Peng}\thanks{Corresponding Author. Email: zhenpeng27@outlook.com.\\ The code and data are available at: https://github.com/yimingxu24/CMUCL}} 
\author[A,B]{\fnms{Bin}~\snm{Shi}} 
\author[A,B]{\fnms{Jiarun}~\snm{Chen}} 
\author[D]{\fnms{Xingbo}~\snm{Fu}} 
\author[D]{\fnms{Song}~\snm{Wang}} 
\author[B,C]{\fnms{Bo}~\snm{Dong}}

\address[A]{School of Computer Science and Technology, Xi'an Jiaotong University}
\address[B]{Shaanxi Provincial Key Laboratory of Big Data Knowledge Engineering, Xi’an Jiaotong University}
\address[C]{School of Distance Education, Xi’an Jiaotong University}
\address[D]{University of Virginia}


\begin{abstract}
The widespread application of graph data in various high-risk scenarios has increased attention to graph anomaly detection (GAD). Faced with real-world graphs that often carry node descriptions in the form of raw text sequences, termed text-attributed graphs (TAGs), existing graph anomaly detection pipelines typically involve shallow embedding techniques to encode such textual information into features, and then rely on complex self-supervised tasks within the graph domain to detect anomalies. However, this text encoding process is separated from the anomaly detection training objective in the graph domain, making it difficult to ensure that the extracted textual features focus on GAD-relevant information, seriously constraining the detection capability. How to seamlessly integrate raw text and graph topology to unleash the vast potential of cross-modal data in TAGs for anomaly detection poses a challenging issue. This paper presents a novel end-to-end paradigm for text-attributed graph anomaly detection, named CMUCL. We simultaneously model data from both text and graph structures, and jointly train text and graph encoders by leveraging cross-modal and uni-modal multi-scale consistency to uncover potential anomaly-related information. Accordingly, we design an anomaly score estimator based on inconsistency mining to derive node-specific anomaly scores. Considering the lack of benchmark datasets tailored for anomaly detection on TAGs, we release 8 datasets to facilitate future research. Extensive evaluations show that CMUCL significantly advances in text-attributed graph anomaly detection, delivering an 11.13\% increase in average accuracy (AP) over the suboptimal.
\end{abstract}

\end{frontmatter}


\section{Introduction}
Graph-structured data is ubiquitous in various security-related applications, which has heightened interest in graph anomaly detection (GAD) within the data mining community~\cite{zheng2024survey,li2025context,xu2025ted}, and substantial efforts have been devoted to detecting abnormal objects that deviate significantly from the patterns of the majority in attributed graphs ~\cite{ma2021comprehensive}. However, beyond conventional attributed graphs, which often describe nodes as numerical multi-dimensional feature vectors, many real-world graphs are associated with raw text sequences and are termed text-attributed graphs (TAGs)~\cite{chen2024exploring}. TAGs are prevalent in multiple scenarios, such as product titles, summaries, and reviews in e-commerce networks~\cite{yan2023comprehensive}, or abstracts associated with each publication in citation networks~\cite{he2023harnessing}. The integration of graph topology with textual attributes provides a rich semantic source of information. Given that the available annotated data is scarce in GAD, this cross-modal information aids in detecting subtle and intricate anomalies across applications, rendering text-attributed graph anomaly detection (TAGAD) both valuable and intriguing.

To encode textual information in graphs, previous GAD works usually use non-contextualized shallow embeddings~\cite{liu2022bond,tang2024gadbench}, such as skip-gram or bag-of-words (BoW). After that, they detect anomalies by meticulously designing various self-supervised strategies to fully utilize unlabeled attributed graph data. Broadly speaking, existing techniques could be categorized into two main classes: reconstruction-based~\cite{ding2019deep,roy2024gad} and contrastive methods~\cite{liu2021anomaly,xu2025revisiting}. Reconstruction-based approaches detect anomalies by identifying significant discrepancies in the graph reconstruction errors, while contrastive methods uncover anomalies by assessing the mismatch between nodes and their neighborhoods. Essentially, both approaches focus on identifying inconsistencies within the graph domain. Despite achieving empirical success, an inherent limitation of the above methods is the disconnection between the textual encoders used in the feature encoding process and the GNN encoders in the GAD task. Specifically, the text encoder parameters are frozen and not involved in training, and some statistical-based methods do not even require pre-training. This disconnection prevents the encoding process of textual features from being specifically guided by the training gradients of the GAD tasks. Consequently, the extracted features are coarse-grained~\cite{wen2023augmenting,he2023harnessing} and fail to direct the text encoder to focus on GAD-relevant information, significantly constraining the detection capability. So, a natural question arises here: \textit{How to effectively design self-supervised tasks to unleash the vast potential of cross-modal data in TAGs to enhance anomaly detection}.

The core of anomaly detection in TAGs lies in seamlessly integrating cross-modal information, specifically raw text sequences and graph topology. From the perspective of expertise in data of each modal, language models (LMs) exhibit profound context-aware knowledge and exceptional semantic understanding. Simultaneously, graph neural networks (GNNs) effectively preserve the intricate topological information of graphs with high fidelity. To synergistically combine the strengths of both GNNs and LMs, a promising approach is to adopt a unified end-to-end training paradigm that jointly models textual attributes and graph topology to tackle the above challenge.
Building upon this architecture, for normal nodes (i.e., most nodes in the graph), the contextual information captured from the textual domain and the topological structure derived from the graph domain fundamentally strive to recover a representation of reality, i.e., a representation of the joint distribution over events in the world that generate the data we observe~\cite{huh2024platonic}, that satisfies consistency principle. In contrast, abnormal objects may not adhere to this normative alignment. Anomalies typically arise when the textual context and structure are inconsistent, resulting in corresponding contextual or structural anomalies~\cite{ma2021comprehensive,liu2021anomaly}. In this sense, we can train a joint modeling framework using cross-modal consistency mining objectives that are highly relevant to TAG anomaly detection. 

This paper presents a novel unsupervised anomaly detection framework tailored for text-attributed graphs named \textbf{CMUCL}, which incorporates \textbf{C}ross-\textbf{M}odal and \textbf{U}ni-modal multi-scale \textbf{C}ontrastive \textbf{L}earning. We focus on TAGs that contain raw textual information, rather than attributed graphs that rely solely on pre-extracted shallow features. Specifically, we first use a GNN-based graph encoder and an LM-based text encoder to learn node-level information from two perspectives in the graph and text domains, respectively. Note that anomalies often occur at different scales~\cite{jin2021anemone}, to capture anomalies at multiple scales, we also encode the context of node neighborhoods using the graph encoder and text encoder in both domains, learning more representative and intrinsic subgraph-level features. 
Building on this, CMUCL trains a joint framework by maximizing the consistency of cross-modal multi-scale contrasts (inner-scale and inter-scale of the same object in different modalities) and uni-modal multi-scale contrasts (graph or text modality node-context contrasts). 
Joint training ensures that features extracted from the text domain are directly anomaly-relevant, leveraging a shared latent space to effectively integrate complementary cues from both text and graph domains.
Finally, we design an anomaly score estimator based on inconsistency mining, which calculates node-specific anomaly scores via positive and negative pairs, along with cross-entropy information from contrastive views. Our main contributions are summarized as follows:

$\bullet$ \textbf{\textit{Foundational Impact}}: To the best of our knowledge, this is the first work to propose the task of text-attributed graph anomaly detection, highlighting that incorporating cross-modal textual information helps unlock the vast potential of self-supervised GAD methods. Building on the success of LMs, this study establishes a foundation for integrating LMs to drive future developments in GAD.

$\bullet$ \textbf{\textit{Novel Algorithm}}: We propose a novel anomaly detection framework that jointly optimizes two modality encoders via cross-modal and uni-modal multi-scale contrastive learning and presents an anomaly score estimator to convert consistency measures into concrete anomaly scores.

$\bullet$ \textbf{\textit{Dataset Contribution}}: To address the current lack of text-attributed graph datasets with anomalies, we release eight novel datasets, including a large-scale dataset with over 1.1 million nodes and 6.3 million edges, to facilitate future TAGAD studies.

$\bullet$ \textbf{\textit{State-of-the-Art Performance}}: Extensive evaluation of CMUCL on eight datasets and eleven baselines shows an average improvement of 4.68\% in AUC and 11.13\% in AP over the suboptimal, highlighting its effectiveness in anomaly detection. \looseness=-1

\section{Related Work}
\subsection{Graph Anomaly Detection}
Early works typically employed non-deep learning paradigms, such as clustering-based~\cite{perozzi2014focused} and matrix factorization techniques~\cite{bandyopadhyay2019outlier}, to identify anomalies in network analysis. The rapid advancement of GNNs has driven the development of deep learning-based GAD methods~\cite{ma2021comprehensive}. DOMINANT~\cite{ding2019deep} measures node anomaly scores based on feature and structure matrices reconstruction errors. GAD-NR~\cite{roy2023gad} introduces reconstructed node neighborhoods to detect anomalies. With the success of contrastive learning, CoLA pioneeringly introduces contrastive learning to GAD. Building on this foundation, ANEMONE~\cite{jin2021anemone} incorporates node-node contrast, while GRADATE~\cite{duan2023graph}, Sub-CR~\cite{zhang2022reconstruction}, and SAMCL~\cite{hu2023samcl} further integrate subgraph-to-subgraph contrast to more accurately estimate node anomaly scores. \cite{liu2024arc,lin2024unigad} attempt to develop a general framework.
Despite significant progress, existing GAD methods focus solely on attributed graphs, overlooking the anomalous cues in textual information, which leads to suboptimal solutions.

\subsection{Graph Contrastive Learning}
Contrastive learning is a significant paradigm in self-supervised learning~\cite{xu2023cldg}, widely favored for its ability to avoid the costs of annotating large-scale datasets~\cite{shi2023edge,xu2024learning}. The core idea involves using contrastive loss to pull the embeddings of matched positive pairs together while pushing the embeddings of non-matched negative pairs apart in the feature space. DGI~\cite{velickovic2019deep} extends contrastive learning to graphs and maximizes the mutual information between global graph embeddings and local node embeddings. GMI~\cite{peng2020graph} maximizes the graphical mutual information between the input and output of a graph neural encode to improve the DGI. MVGRL~\cite{hassani2020contrastive} introduces graph diffusion~\cite{gasteiger2019diffusion} to create additional graph views for contrast.

\subsection{Language Models on Graphs}
The remarkable achievements of language models (LMs) across various domains have increasingly drawn the attention of graph machine learning researchers~\cite{li2023survey,jin2023large}. Based on the role LMs play in the graph learning pipeline, they can be categorized as enhancers, predictors, or aligners. GIANT~\cite{chien2021node} and TAPE~\cite{he2023harnessing} use LMs as enhancers to enrich the node-related semantic knowledge to improve the supervised node classification performance of GNN. LLaGA~\cite{chen2024llaga} employs templates to transform graph structure into sequences and utilize LMs as a predictor to perform classification. GLEM~\cite{zhao2022learning} uses an EM framework to integrate GNNs and LLMs, with each model iteratively generating pseudo-labels for the other. 
However, most of these methods are semi-supervised, multi-stage, and primarily focus on node classification tasks. Overall, the integration of LMs and graphs for anomaly detection remains in its infancy and requires further exploration.

\section{Problem Formulation}
In this paper, we focus on the task of self-supervised text-attributed graph anomaly detection, formalized as follows:

\begin{definition}[Text-Attributed Graph]
Given a text-attributed graph (TAG) $\mathcal{G}= \left ( \mathcal{V},\mathcal{E}, \mathcal{T},\mathbf{A} \right )$, where $\mathcal{V}=\left\{ v_{1},\cdots ,v_{n}\right\}$ is the set of $n$ nodes paired with raw textual information $\mathcal{T}=\left\{ t_{1},\cdots ,t_{n}\right\}$, and $t_{i}\in \mathcal{D}^{L_i}$ with $\mathcal{D}$ is the words or tokens dictionary, $L_i$ as the sequence length of node $i$. $\mathcal{E}$ is a set of edges and $\mathbf{A}\in \mathbb{R}^{n\times n}$ is the adjacency matrix, where $\mathbf{A}\left [ i,j \right ]=1$ indicates an edge between node $i$ and $j$, and $\mathbf{A}[i,j] = 0$ means not directly connected.
\end{definition}
 
\begin{definition}[Text-Attributed Graph Anomaly Detection]
Given a TAG $\mathcal{G}= \left ( \mathcal{V},\mathcal{E}, \mathcal{T},\mathbf{A} \right )$, the goal of TAGAD is to learn an anomaly score function $f: \mathcal{V} \rightarrow \mathbb{R}$ using a self-supervised approach that estimates an anomaly score $f(v_i)$ to each node $v_i \in \mathcal{V}$, where higher scores indicate higher likelihoods of being anomalous.
\end{definition}

\begin{figure*}[!htb]
  \centering
  \includegraphics[width=0.9\linewidth]{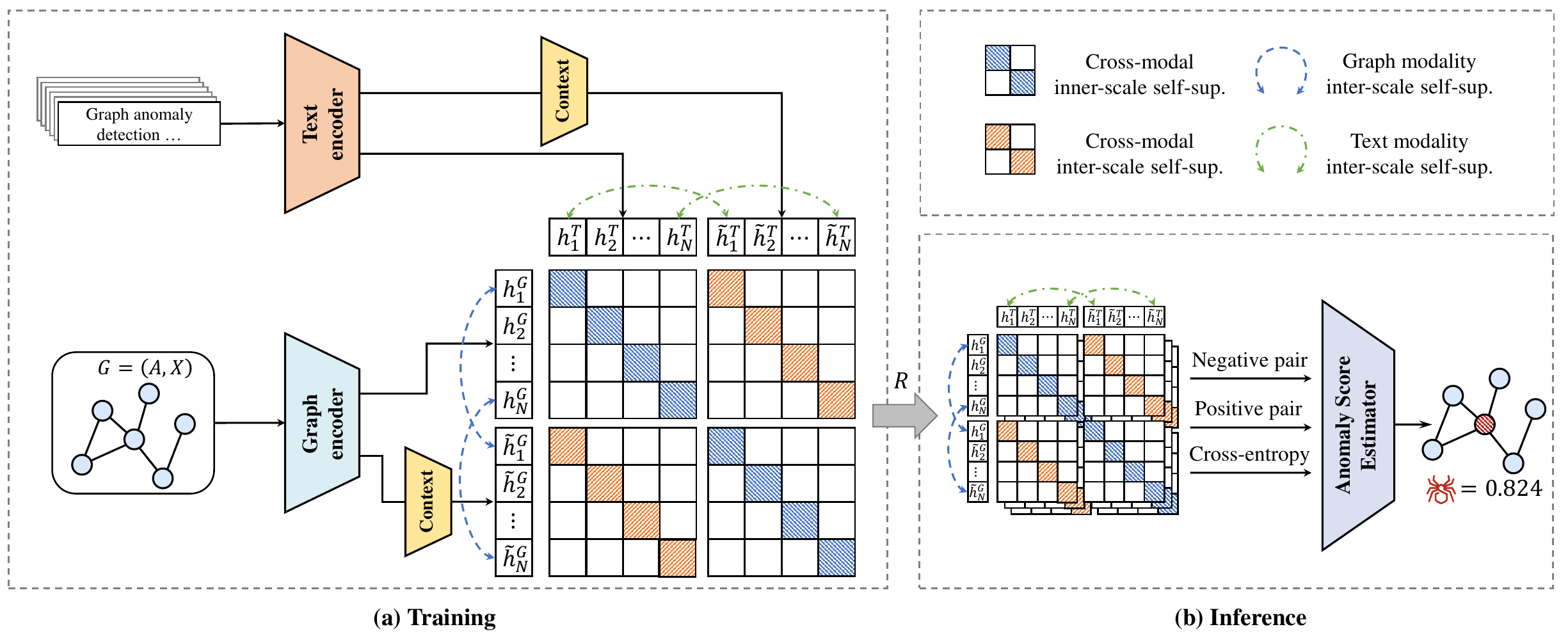}
  \caption{The overall pipeline of CMUCL.}
  \label{fig:overview}
\end{figure*}

\section{Methodology}
In this section, we present an overview of the CMUCL framework, as illustrated in Figure~\ref{fig:overview}. First, we devise the bi-modal feature extraction process. Next, we provide cross-modal and uni-modal multi-scale self-supervised learning approaches. Finally, we outline the training objective during the training phase and the estimator that converts consistency measures into concrete anomaly scores during the inference phase.

\subsection{Bi-Modal Feature Extraction}
Existing GAD methods often rely on pre-extracted shallow features, which fail to provide the comprehensive representation necessary for effective anomaly detection. To address these limitations, we jointly optimize both the text and graph domains to learn a bi-modal embedding space, thereby unleashing the full potential of the data. 
\paragraph{Node-Level Embeddings.}
For the text domain, we use a Transformer encoder~\cite{vaswani2017attention} to capture the contextual information of the text attributes associated with each node. Given a text attribute $t_i$ of node $v_i$, the encoded representation $\mathbf{h}_i^{T}$ is obtained from the end-of-sentence (EOS) token:
\begin{equation}
\mathbf{h}_i^{T} = \text{Transformer}(t_i).
\end{equation}

For the graph domain, we use a graph convolutional network (GCN)~\cite{kipf2016semi} to capture the topological structure of the graph. Given a node $v_i$ and its neighborhood $\mathcal{N}(i)$, the encoded representation $\mathbf{h}_i^{G}$ can be formally described as:
\begin{equation}
\mathbf{h}_i^{G} = \text{GNN}(v_i, \{v_j | j \in \mathcal{N}(i)\}).
\end{equation}

\paragraph{Context-Level Embeddings.}
In addition to node-level embeddings, we also focus on capturing context-level information to detect anomalous patterns at different scales. This involves encoding the summary of node neighborhoods using both the graph and text embeddings. The context-level embeddings are computed via a readout module:
\begin{equation}
\begin{aligned}
\mathbf{\widetilde{h}}_{i}^{T} &= \text{Readout}(\mathbf{H}_{i}^{T}) = \sum_{k\in \mathcal{N}(i)}\frac{\mathbf{h}_k^{T}}{\left | \mathcal{N}(i)\right |},\\
\mathbf{\widetilde{h}}_{i}^{G} &= \text{Readout}(\mathbf{H}_{i}^{G}) = \sum_{k\in \mathcal{N}(i)}\frac{\mathbf{h}_k^{G}}{\left | \mathcal{N}(i)\right |},
\end{aligned}
\end{equation}
where $\mathbf{H}_{i}^{T}$ and $\mathbf{H}_{i}^{G}$ are the neighbor feature matrices for node $i$ in the text and graph domains, respectively.  
$\mathcal{N}(i)$ is the number of neighbors of node $i$. $\mathbf{\widetilde{h}}_{i}^{T}$ and $\mathbf{\widetilde{h}}_{i}^{G}$ are the context embeddings of the text domain and graph domain of node $i$. 

\subsection{Cross-Modal Multi-scale Contrast}
Given a text-attributed graph $\mathcal{G}$, after bi-model feature extraction, the text encoder maps the text attribute and context information to a normalized representation $\mathbf{h}_i^{T}$ and $\mathbf{\widetilde{h}}_{i}^{T}$. Similarly, the graph encoder maps the structural and context information of each node $i$ to a normalized representation $\mathbf{h}_i^{G}$ and $\mathbf{\widetilde{h}}_{i}^{G}$. Essentially, both the text domain and the graph domain are abstractions of real-world entities, and the bi-modal encoders aim to reconstruct representations of reality, i.e., representations of the joint distribution of events in the world that generate the data we observe. Therefore, the goal of cross-modal contrastive learning is to learn the cross-modal consistency between textual and graph structures for normal nodes. To achieve this, we design a cross-modal multi-scale contrastive learning framework that captures the intrinsic relationships between node representations across different modalities and scales. We define two types of contrasts. \looseness=-1

\paragraph{Cross-Modal Inner-Scale Self-Supervision.}
This contrast aims to align node-level or context-level representations between the text and graph domains. For each node $i$, we maximize the consistency between its text node-level representations $\mathbf{h}_i^{T}$ and its graph node-level representations $\mathbf{h}_i^{G}$, as well as the consistency between text context-level representations $\mathbf{\widetilde{h}}_{i}^{T}$ and graph context-level representations $\mathbf{\widetilde{h}}_{i}^{G}$. Simultaneously, we minimize the consistency between node-level and context-level representations of other nodes in the training batch. This ensures the embeddings capture the consistent patterns shared between the text and graph modalities for the same node. Then the node-level infoNCE~\cite{van2018representation} loss can be expressed as:
\begin{equation}
\mathcal{L}_{\text{g2t}}^{\text{nn}} = -\frac{1}{N} \sum_{i=1}^{N} \log \frac{\exp\left(\text{sim}(\mathbf{h}_i^{G}, \mathbf{h}_i^{T})/\tau \right)}{\sum_{j=1}^{N} \exp\left(\text{sim}(\mathbf{h}_i^{G}, \mathbf{h}_j^{T})/\tau \right)},
\end{equation}
where $\text{sim}(\cdot, \cdot)$ is the similarity between two representations (such as dot product or cosine similarity). $\tau$ is a temperature parameter and $N$ is the batch size. Additionally, the context-level loss function is defined as follows:
\begin{equation}
\mathcal{L}_{\text{g2t}}^{\text{cc}} = -\frac{1}{N} \sum_{i=1}^{N} \log \frac{\exp\left(\text{sim}(\mathbf{\widetilde{h}}_{i}^{G}, \mathbf{\widetilde{h}}_{i}^{T})/\tau \right)}{\sum_{j=1}^{N} \exp\left(\text{sim}(\mathbf{\widetilde{h}}_{i}^{G}, \mathbf{\widetilde{h}}_{j}^{T})/\tau \right)}.
\end{equation}

The cross-modal inner-scale loss is formulated as:
\begin{equation}
\mathcal{L}_{\text{g2t}}^{\text{inner}} = \mathcal{L}_{\text{g2t}}^{\text{nn}} + \mathcal{L}_{\text{g2t}}^{\text{cc}}.
\end{equation}

\paragraph{Cross-Modal Inter-Scale Self-Supervision.}
Since anomalies can manifest at different granular levels~\cite{jin2021anemone}, capturing these discrepancies requires robust inter-scale alignment. Therefore, we extend the alignment mechanism to different scales, ensuring that the patterns of normal nodes are consistent not only within the same scale across modalities but also across multiple scales. In other words, maintain consistency at the node-level and context-level across modalities. By incorporating inter-scale supervision, we aim to leverage the rich multi-scale relationships inherent in the data, further enhancing the model's ability to detect anomalies. The cross-modal inner-scale loss is expressed as: \looseness=-1
\begin{equation}
\mathcal{L}_{\text{g2t}}^{\text{inter}} = \mathcal{L}_{\text{g2t}}^{\text{nc}} + \mathcal{L}_{\text{g2t}}^{\text{cn}},
\end{equation}
where $\mathcal{L}_{\text{g2t}}^{\text{nc}}$ and $\mathcal{L}_{\text{g2t}}^{\text{cn}}$ denote the alignment between the node-level/context-level features of the graph domain and the context-level/node-level features of the text domain, respectively. The formulas for both are defined as follows:
\begin{align}
\mathcal{L}_{\text{g2t}}^{\text{nc}} &= -\frac{1}{N} \sum_{i=1}^{N} \log \frac{\exp\left(\text{sim}(\mathbf{h}_i^{G}, \mathbf{\widetilde{h}}_{i}^{T})/\tau \right)}{\sum_{j=1}^{N} \exp\left(\text{sim}(\mathbf{h}_i^{G}, \mathbf{\widetilde{h}}_{j}^{T})/\tau \right)}, \\
\mathcal{L}_{\text{g2t}}^{\text{cn}} &= -\frac{1}{N} \sum_{i=1}^{N} \log \frac{\exp\left(\text{sim}(\mathbf{\widetilde{h}}_{i}^{G}, \mathbf{h}_i^{T})/\tau \right)}{\sum_{j=1}^{N} \exp\left(\text{sim}(\mathbf{\widetilde{h}}_{i}^{G}, \mathbf{h}_j^{T})/\tau \right)}.
\end{align}

The final cross-modal contrast loss function is:
\begin{equation}
\mathcal{L}^{\text{cross}} = \mathcal{L}_{\text{g2t}}^{\text{inner}} + \mathcal{L}_{\text{g2t}}^{\text{inter}},
\end{equation}

\subsection{Uni-Modal Multi-scale Contrast}
Ensuring alignment at different scales within the graph modality has been a central focus of existing works, such as the GAD method~\cite{duan2023graph}, and is key to their success. Therefore, after establishing the cross-modal alignment framework, it is crucial to ensure the consistency of representations at different scales within each individual modality, whether it be the text domain or the graph domain. The goal of uni-modal multi-scale contrastive learning is to capture the inherent relationships within a single modality by aligning representations at various scales. The uni-modal contrast loss function is given as:
\begin{equation}
\begin{aligned}
\mathcal{L}^{\text{uni}} &= \mathcal{L}_{\text{t2t}}^{\text{inter}} + \mathcal{L}_{\text{g2g}}^{\text{inter}} \\
&= -\frac{1}{N} \sum_{i=1}^{N} \left[ \log \frac{\exp\left(\text{sim}(\mathbf{h}_i^{T}, \mathbf{\widetilde{h}}_{i}^{T})/\tau \right)}{\sum_{j=1}^{N} \exp\left(\text{sim}(\mathbf{h}_i^{T}, \mathbf{\widetilde{h}}_{j}^{T})/\tau \right)} \right. \\
&\quad \left. + \log \frac{\exp\left(\text{sim}(\mathbf{h}_i^{G}, \mathbf{\widetilde{h}}_{i}^{G})/\tau \right)}{\sum_{j=1}^{N} \exp\left(\text{sim}(\mathbf{h}_i^{G}, \mathbf{\widetilde{h}}_{j}^{G})/\tau \right)} \right],
\end{aligned}
\end{equation}
where $\mathcal{L}_{\text{t2t}}^{\text{inter}}$ and $\mathcal{L}_{\text{g2g}}^{\text{inter}}$ involve the alignment of node-level and context-level representations within the text and graph modalities, respectively.

\subsection{Training Objective}
By incorporating both cross-modal and uni-modal multi-scale contrastive learning, we optimize the joint objective function:
\begin{equation}
\label{eq:loss}
\mathcal{L} = \mathcal{L}^{\text{cross}} +  \gamma \mathcal{L}^{\text{uni}},
\end{equation}
where $\gamma$ is a tradeoff parameter that balances the importance between cross-modal and uni-modal contrast.

\subsection{Anomaly Score Estimator}
After effective self-supervised training, we introduce an anomaly score estimator based on inconsistency mining, to convert cross-modal node representations into specific node anomaly scores during the inference phase. The anomaly score is computed based on each contrastive view from the training phase and then combined to form the final score. In each view, normal nodes should be similar to their positive pair and dissimilar to their negative pairs, and the cross-entropy loss should be minimized. In contrast, abnormal objects typically do not satisfy the above consistency. To capture the inconsistency, we integrate these three components to compute the anomaly score for each contrastive view. The anomaly score for node $i$ is as follows:
\begin{equation}
\label{eq:ase}
s_i = \sum_{w}^{W} \gamma_i \left( s_{i,w}^n - s_{i,w}^p + \mathcal{C}_{i,w} \right),
\end{equation}
where $W$ represents all contrastive views used in the training phase. Here we take $w$ to represent cross-modal node-level contrastive views as an example, the positive similarity $s_{i,v}^p = \text{sim}(\mathbf{h}_i^{G}, \mathbf{h}_i^{T})/\tau$, the negative similarity $s_{i,v}^n = \frac{\sum_{j\neq i}^{N} \left(\text{sim}(\mathbf{h}_i^{G}, \mathbf{h}_j^{T})/\tau \right) }{N-1}$, and cross-entropy loss for node $i$ is $\mathcal{C}_{i,v} = - \log \frac{\exp\left(\text{sim}(\mathbf{h}_i^{G}, \mathbf{h}_i^{T})/\tau \right)}{\sum_{j=1}^{N} \exp\left(\text{sim}(\mathbf{h}_i^{G}, \mathbf{h}_j^{T})/\tau \right)}$. $\gamma_i$ takes the same values as in Eq.~\eqref{eq:loss}.

Due to the limited number of negative samples in a batch, which may not provide enough discriminative power, we perform multiple rounds of detection. Anomalous nodes usually show higher scores and instability across batches. The final anomaly score is determined by assessing consistency and stability:
\begin{align}
\overline{S}_{i} &= \frac{\sum_{r=1}^{R} s_{i}^{r}}{R}, \label{eq:cons} \\
S_i &= \overline{S_{i}} + \sqrt{\frac{\sum_{r=1}^{R} \left(s_i^r - \overline{S}_i \right)^{2}}{R}}, \label{eq:estimator}
\end{align}
where $R$ is the number of sampling rounds.

\textbf{Discussion}. Existing GAD methods predominantly focus on designing increasingly complex self-supervised tasks for attributed graphs, neglecting the rich textual information often present in graphs. This work is the first to highlight the critical role of cross-modal information for anomaly detection, an aspect previously unexplored in the literature. Similar to leveraging graph structure to improve early anomaly detection (AD) methods that relied solely on attribute features, we further extend GAD to TAGAD by integrating textual data. Our proposed cross-modal multi-scale consistency objective enables the text encoder to capture GAD-relevant patterns effectively, exploiting the complementary strengths of graph and textual domains to train a unified framework that enhances anomaly detection performance. Building on the success of LMs, this study lays the foundation for more advanced methodologies and evaluations, demonstrating the potential of leveraging LMs to propel future progress in GAD.

\begin{table*}[ht]\small
\centering
\caption{Experimental results for anomaly detection on eight datasets (OOM: CPU/CUDA Out of Memory). We report both mean AUC and AP. \textbf{Bold} represents the optimal in the unsupervised method, and  \underline{underlined} represents the runner-up.}
\resizebox{1\textwidth}{!}{
\begin{tabular}{ccccccccc}
\toprule[1pt]
\multirow{2}{*}{Method} & \multicolumn{2}{c}{Citeseer} & \multicolumn{2}{c}{Pubmed} & \multicolumn{2}{c}{History} & \multicolumn{2}{c}{Photo} \\
\cline{2-9}
 & AUC & AP & AUC & AP & AUC & AP & AUC & AP \\
\hline
LOF
& 52.51$_{\pm 0.00}$  & 4.12$_{\pm 0.00}$ 
& 64.10$_{\pm 0.00}$  & 8.39$_{\pm 0.00}$ 
& 55.71$_{\pm 0.00}$  & 5.22$_{\pm 0.00}$ 
& 53.30$_{\pm 0.00}$  & 4.30$_{\pm 0.00}$  \\

SCAN
& 72.61$_{\pm 0.91}$  & 17.25$_{\pm 0.49}$ 
& 60.16$_{\pm 0.52}$  & 10.41$_{\pm 0.64}$ 
& 57.27$_{\pm 0.89}$  & 5.00$_{\pm 0.70}$ 
& 55.66$_{\pm 0.46}$  & 4.56$_{\pm 0.51}$  \\

Radar
& 42.78$_{\pm 0.30}$  & 3.47$_{\pm 0.50}$ 
& 51.88$_{\pm 0.35}$  & 3.92$_{\pm 0.24}$ 
& 48.48$_{\pm 1.05}$  & 3.60$_{\pm 0.76}$ 
& 49.92$_{\pm 0.43}$  & 3.81$_{\pm 0.53}$  \\
\hline

AEGIS
& 52.29$_{\pm 1.32}$  & 4.15$_{\pm 0.51}$ 
& 49.08$_{\pm 1.10}$  & 3.66$_{\pm 0.50}$ 
& 52.28$_{\pm 1.97}$  & 4.23$_{\pm 0.75}$ 
& 53.42$_{\pm 1.91}$  & 4.32$_{\pm 0.54}$  \\

MLPAE
& 50.99$_{\pm 0.55}$  & 4.03$_{\pm 0.10}$ 
& 50.56$_{\pm 0.26}$  & 3.93$_{\pm 0.39}$ 
& 47.54$_{\pm 0.16}$  & 3.67$_{\pm 0.53}$ 
& 47.48$_{\pm 0.87}$  & 3.72$_{\pm 0.58}$  \\

DOMINANT
& 61.10$_{\pm 0.37}$  & 6.74$_{\pm 0.05}$ 
& 67.32$_{\pm 0.61}$  & 7.54$_{\pm 0.81}$ 
& 63.46$_{\pm 0.23}$  & 6.17$_{\pm 0.15}$ 
& 60.10$_{\pm 0.52}$  & 5.73$_{\pm 0.16}$  \\

GAD-NR
& $\underline{76.74_{\pm 0.88}}$  & 32.01$_{\pm 0.27}$ 
& 66.25$_{\pm 0.44}$  & 6.34$_{\pm 0.04}$ 
& 65.86$_{\pm 0.13}$  & 5.89$_{\pm 0.22}$ 
& 63.50$_{\pm 0.01}$  & 5.75$_{\pm 0.03}$  \\
\hline

CoLA
& 73.68$_{\pm 0.93}$  & 31.67$_{\pm 0.73}$ 
& $\underline{78.83_{\pm 0.55}}$  & 18.93$_{\pm 0.62}$ 
& 78.72$_{\pm 0.09}$  & 26.48$_{\pm 0.55}$ 
& 70.47$_{\pm 0.08}$  & 8.64$_{\pm 0.72}$  \\

ANEMONE
& 74.45$_{\pm 0.19}$  & 31.78$_{\pm 0.66}$ 
& 78.81$_{\pm 0.02}$  & $\underline{19.43_{\pm 0.54}}$ 
& $\underline{79.14_{\pm 0.02}}$  & $\underline{27.20_{\pm 0.66}}$ 
& $\underline{71.97_{\pm 0.07}}$  & $\underline{11.28_{\pm 0.85}}$  \\

SL-GAD
& 72.51$_{\pm 1.60}$  & 34.34$_{\pm 0.08}$ 
& 75.19$_{\pm 0.35}$  & 16.71$_{\pm 0.05}$ 
& 74.08$_{\pm 0.28}$  & 20.22$_{\pm 0.41}$ 
& 62.78$_{\pm 0.27}$  & 6.89$_{\pm 0.75}$  \\

GRADATE
& 73.66$_{\pm 0.42}$  & $\mathbf{35.39}_{\pm 0.66}$ 
& 67.49$_{\pm 0.09}$  & 12.13$_{\pm 0.18}$ 
& 73.25$_{\pm 0.10}$  & 14.50$_{\pm 0.55}$ 
& 65.79$_{\pm 0.10}$  & 10.26$_{\pm 0.25}$  \\

\hline
\rowcolor[HTML]{E9E9E9}
\textbf{CMUCL}
& $\mathbf{82.29}_{\pm 1.25}$  & $\underline{34.62_{\pm 0.24}}$ 
& $\mathbf{81.78}_{\pm 0.27}$  & $\mathbf{30.96}_{\pm 0.15}$ 
& $\mathbf{79.69}_{\pm 0.31}$  & $\mathbf{36.80}_{\pm 0.66}$ 
& $\mathbf{74.26}_{\pm 0.17}$  & $\mathbf{22.90}_{\pm 0.21}$  \\
\end{tabular}}


\resizebox{1\textwidth}{!}{
\begin{tabular}{ccccccccc}
\toprule[1pt]
\multirow{2}{*}{Method} & \multicolumn{2}{c}{Computers} & \multicolumn{2}{c}{Children} & \multicolumn{2}{c}{ogbn-Arxiv} & \multicolumn{2}{c}{CitationV8} \\
\cline{2-9}
 & AUC & AP & AUC & AP & AUC & AP & AUC & AP \\
\hline
LOF
& 54.80$_{\pm 0.00}$  & 4.43$_{\pm 0.00}$ 
& 56.37$_{\pm 0.00}$  & 5.21$_{\pm 0.00}$ 
& 71.85$_{\pm 0.00}$  & 19.53$_{\pm 0.00}$ 
& $\underline{63.01_{\pm 0.00}}$  & 7.10$_{\pm 0.00}$  \\

SCAN
& 55.72$_{\pm 0.41}$  & 4.57$_{\pm 0.36}$ 
& 54.58$_{\pm 0.70}$  & 4.43$_{\pm 0.36}$ 
& 58.98$_{\pm 0.81}$  & 5.48$_{\pm 0.73}$ 
& 62.33$_{\pm 0.46}$  & $\underline{20.92_{\pm 0.72}}$  \\

Radar
& 49.66$_{\pm 0.13}$  & 3.82$_{\pm 0.20}$ 
& 49.24$_{\pm 0.50}$  & 3.69$_{\pm 0.75}$ 
& OOM  & OOM 
& OOM  & OOM  \\
\hline

AEGIS
& 51.46$_{\pm 1.83}$  & 4.01$_{\pm 0.84}$ 
& 51.25$_{\pm 1.43}$  & 3.93$_{\pm 0.64}$ 
& 54.03$_{\pm 1.44}$  & 4.41$_{\pm 0.60}$ 
& OOM  & OOM  \\

MLPAE
& 47.16$_{\pm 0.27}$  & 3.60$_{\pm 0.87}$ 
& 50.18$_{\pm 1.46}$  & 4.00$_{\pm 0.40}$ 
& 51.17$_{\pm 0.50}$  & 3.91$_{\pm 0.23}$ 
& 52.00$_{\pm 0.77}$  & 4.10$_{\pm 0.23}$  \\

DOMINANT
& OOM  & OOM 
& OOM  & OOM 
& OOM  & OOM 
& OOM  & OOM  \\

GAD-NR
& 58.84$_{\pm 0.05}$  & 3.90$_{\pm 0.76}$ 
& 50.56$_{\pm 0.20}$  & 4.91$_{\pm 0.48}$ 
& 65.23$_{\pm 0.51}$  & 6.86$_{\pm 0.15}$ 
& 61.84$_{\pm 0.11}$  & 6.56$_{\pm 0.10}$  \\
\hline

CoLA
& 71.05$_{\pm 0.01}$  & 9.53$_{\pm 0.32}$ 
& 73.70$_{\pm 0.19}$  & 15.09$_{\pm 0.80}$ 
& 81.03$_{\pm 0.15}$  & $\underline{27.73_{\pm 0.40}}$ 
& OOM  & OOM  \\

ANEMONE
& $\underline{72.59_{\pm 0.03}}$  & $\underline{12.33_{\pm 0.16}}$ 
& $\underline{74.29_{\pm 0.10}}$  & $\underline{15.24_{\pm 0.64}}$ 
& 80.97$_{\pm 0.03}$  & 27.49$_{\pm 0.75}$ 
& OOM  & OOM  \\

SL-GAD
& 65.36$_{\pm 0.36}$  & 9.40$_{\pm 0.58}$ 
& 69.86$_{\pm 0.14}$  & 13.33$_{\pm 0.10}$ 
& $\underline{81.16_{\pm 0.21}}$  & 27.64$_{\pm 0.41}$ 
& OOM  & OOM  \\

GRADATE
& OOM  & OOM 
& OOM  & OOM 
& OOM  & OOM 
& OOM  & OOM  \\

\hline
\rowcolor[HTML]{E9E9E9}
\textbf{CMUCL}
& $\mathbf{74.25}_{\pm 0.14}$  & $\mathbf{27.27}_{\pm 0.52}$ 
& $\mathbf{75.08}_{\pm 0.21}$  & $\mathbf{23.21}_{\pm 0.05}$ 
& $\mathbf{84.69}_{\pm 0.20}$  & $\mathbf{40.34}_{\pm 0.26}$ 
& $\mathbf{83.16}_{\pm 0.07}$  & $\mathbf{42.43}_{\pm 0.40}$  \\
\bottomrule[1pt]
\end{tabular}}
\label{tab:ad}
\end{table*}

\begin{table}[t!]
\renewcommand\arraystretch{1.24}
\centering
  \caption{Statistics of the Datasets.}
  \label{tab:SD}
\resizebox{1\hsize}{!}{
  \begin{tabular}{c|ccccc}
    \toprule
    Dataset & Nodes & Edges &Avg. doc length & Attributes & Anomalies \\
    \midrule
    Citeseer & 3,186 & 3,432 & 153.94 & 768 & 128 \\
    Pubmed & 19,717 & 90,368 & 256.08 & 768 & 788 \\
    History & 41,551 & 369,252 & 228.36 & 768 & 1,662 \\
    Photo & 48,362 & 512,933 & 150.25 & 768 & 1,934 \\
    Computers & 87,229 & 742,792 & 93.16 & 768 & 3,490 \\
    Children & 76,875 & 1,574,664 & 209.12 & 768 & 3,076 \\
    ogbn-Arxiv & 169,343 & 1,210,112 & 179.70 & 768 & 6,774 \\
    CitationV8 & 1,106,759 & 6,396,265 & 148.77 & 768 & 44,270 \\
    \bottomrule
  \end{tabular}}
\end{table}

\section{Experiments}
\subsection{Experiment Settings}

\paragraph{Datasets.}
We release eight datasets to facilitate text-attributed graph anomaly detection research. These datasets are categorized into two groups: 1) Citation networks: Citeseer, Pubmed~\cite{chen2024exploring}, ogbn-Arxiv and CitationV8~\cite{yan2023comprehensive}. 2) E-commerce networks: History, Children, Photo and Computers~\cite{yan2023comprehensive}. The statistics of these datasets are demonstrated in Table~\ref{tab:SD}. 
Since existing GAD methods do not handle text, we use LM-based BGE~\cite{bge_embedding} to encode raw text as initial node features for all baselines and the graph domain in CMUCL to ensure fairness. Detailed descriptions are available in the Appendix.

\paragraph{Baselines.} 
For our comparative analysis, we employed a variety of methodologies: density-based model LOF~\cite{breunig2000lof}; structural clustering-based model SCAN~\cite{xu2007scan}; matrix factorization-based model Radar~\cite{li2017radar}; generative adversarial learning-based model AEGIS~\cite{ding2021inductive}; reconstruction-based models MLPAE~\cite{sakurada2014anomaly}, DOMINANT~\cite{ding2019deep} and GAD-NR~\cite{roy2024gad}; contrastive learning-based models CoLA~\cite{liu2021anomaly}, AENMONE~\cite{jin2021anemone}, SL-GAD~\cite{zheng2021generative}, and GRADATE~\cite{duan2023graph}.

\paragraph{Implementation Details.} 
We use a transformer~\cite{vaswani2017attention} as the text encoder, consisting of 12 layers, each with a width of 512, and an 8-head attention mechanism. The graph encoder is a two-layer GCN~\cite{kipf2016semi} with residual connections and a hidden layer dimension of 128. Node features are uniformly mapped to 128 dimensions for alignment. The temperature coefficient $\tau$ is set to 0.07, and $R$ is defined as 256. More details can be found in the Appendix.

\begin{figure*}
    \centering
    \setlength{\belowcaptionskip}{0.15cm}
    \subfloat[Citeseer]{\label{fig:main_roc1}\includegraphics[width=0.25\linewidth]{
    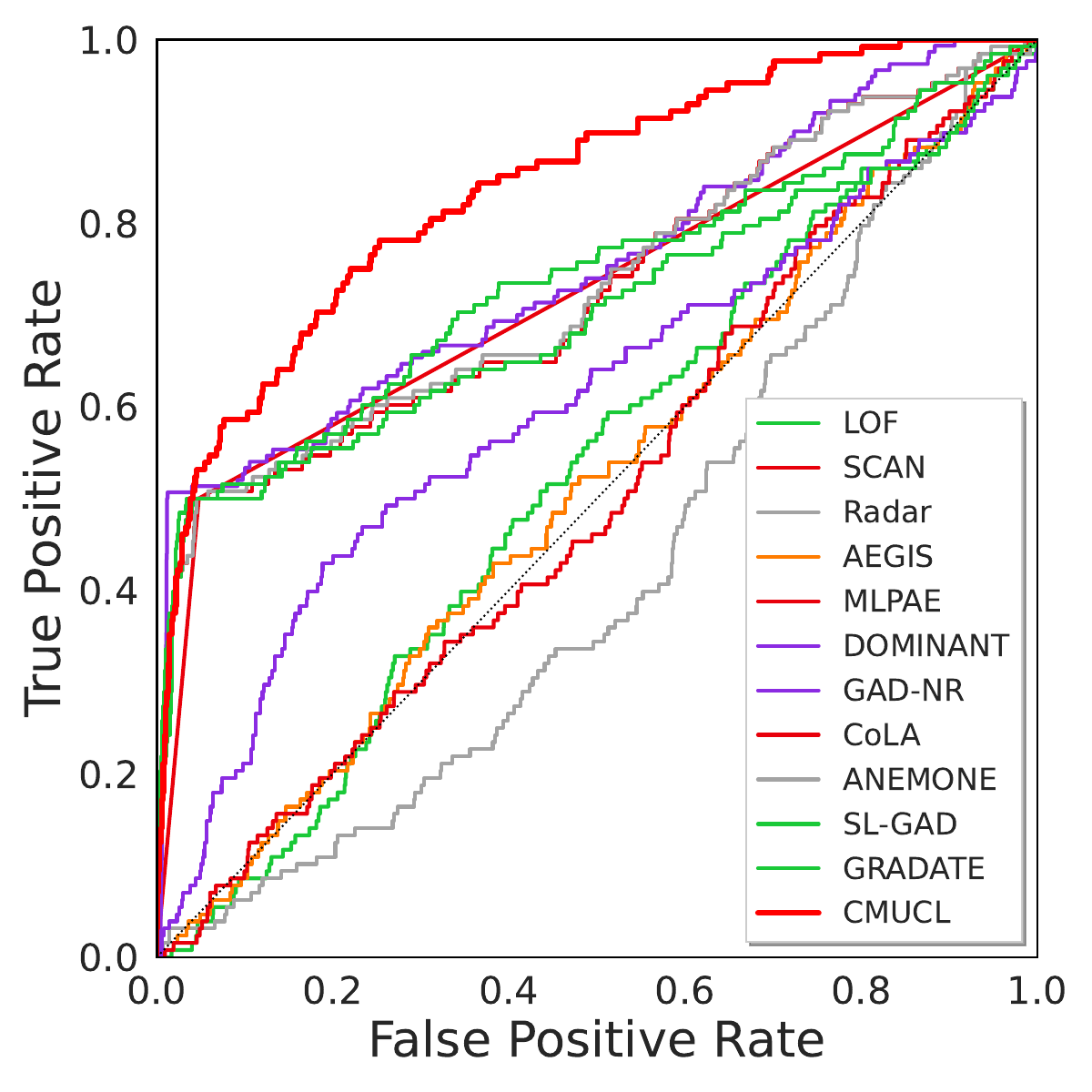}}
    \subfloat[Pubmed]{\label{fig:main_roc2}\includegraphics[width=0.25\linewidth]{
    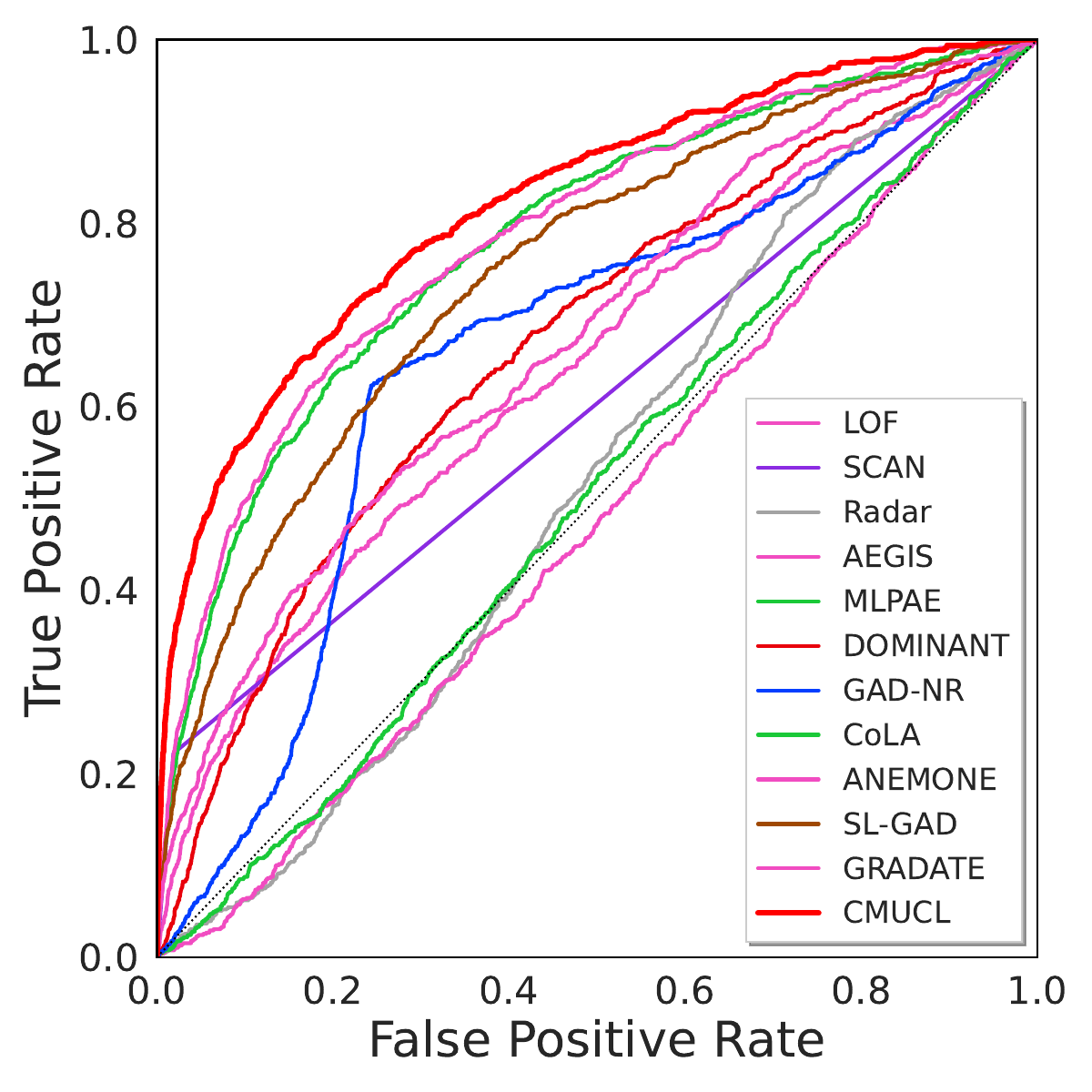}}
    \subfloat[ogbn-Arxiv]{\label{fig:main_roc7}\includegraphics[width=0.25\linewidth]{
    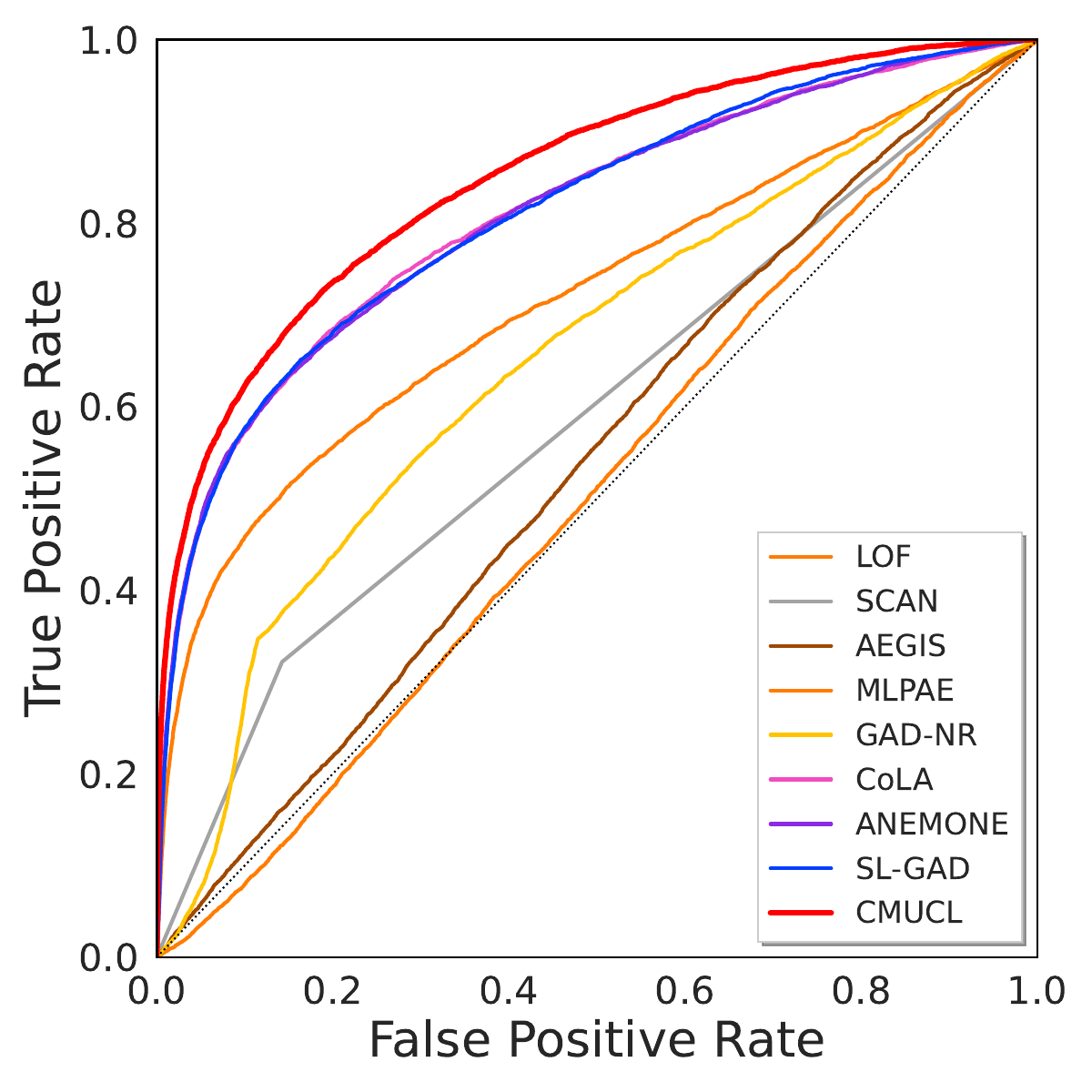}}
    \subfloat[CitationV8]{\label{fig:main_roc8}\includegraphics[width=0.25\linewidth]{
    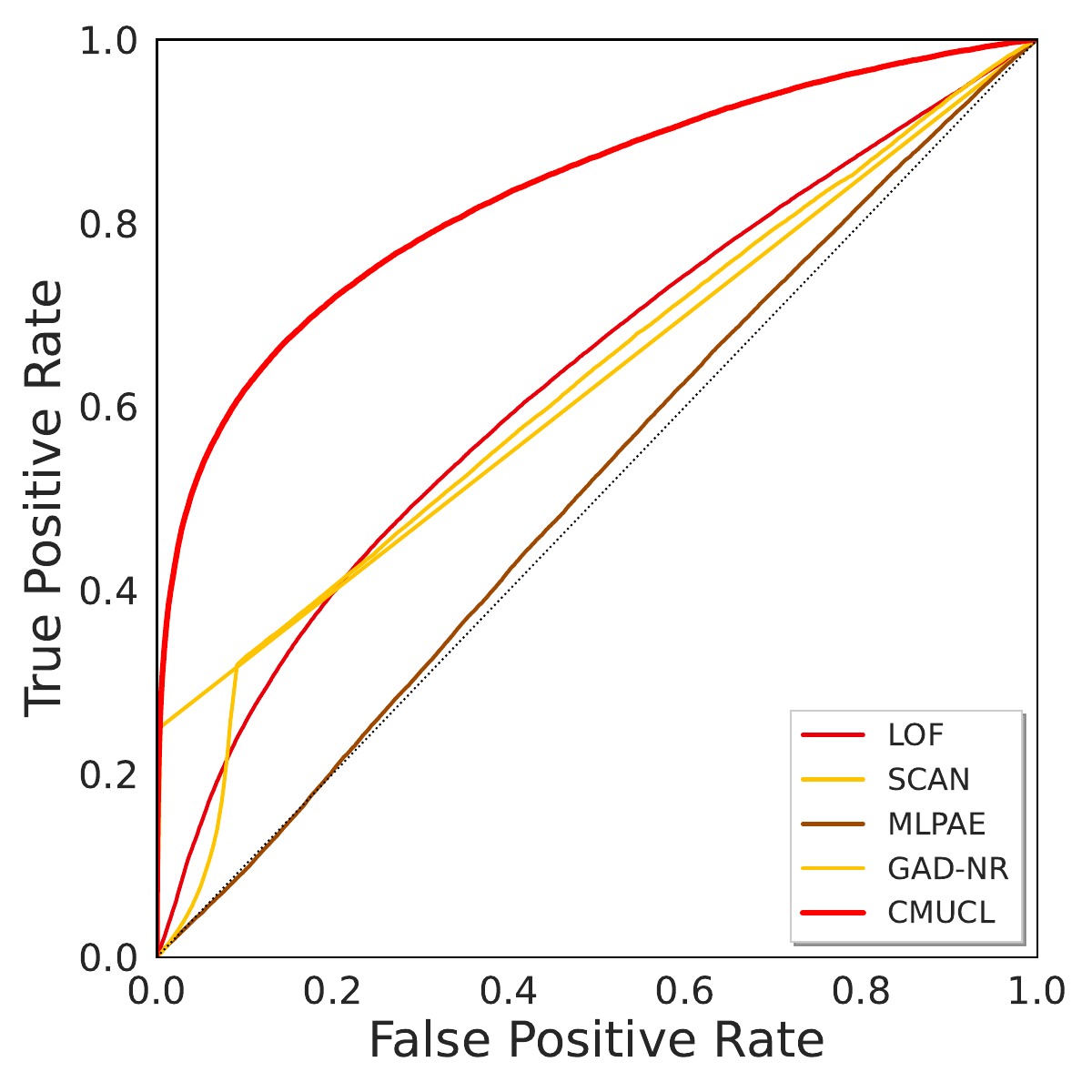}}
    \caption{ROC curves compared on four datasets. A larger area under the curve means better performance. The black dotted lines show the performance of random guessing.}
    \label{fig:main_roc}    
\end{figure*}

\begin{table*}\small
\centering
\renewcommand\arraystretch{1.2}
\caption{Ablation study for different model variants. Bold represents the global optimal.}
\resizebox{\textwidth}{!}{
\begin{tabular}{c|cccccccc}
\toprule
\multicolumn{1}{c|}{Method} & \multicolumn{1}{c}{Citeseer} & \multicolumn{1}{c}{Pubmed} & \multicolumn{1}{c}{History} & \multicolumn{1}{c}{Photo} & \multicolumn{1}{c}{Computers} & \multicolumn{1}{c}{Children} & \multicolumn{1}{c}{ogbn-Arxiv} & \multicolumn{1}{c}{Avg.} \\ \hline
\textrm{w/ cross inner} & $72.91_{\pm 0.30}$ & $76.25_{\pm 0.90}$ & $68.31_{\pm 0.31}$ & $63.68_{\pm 0.42}$ & $66.87_{\pm 0.44}$ & $64.58_{\pm 0.44}$ & $72.69_{\pm 0.30}$ & $69.33_{\pm 0.79} \left ( \downarrow_{9.53} \right )$ \\
\textrm{w/ cross inter} & $81.82_{\pm 1.80}$ & $80.74_{\pm 0.39}$ & $79.25_{\pm 0.28}$ & $73.98_{\pm 0.24}$ & $\textbf{75.10}_{\pm 0.14}$ & $74.13_{\pm 0.13}$ & $82.80_{\pm 0.11}$ & $78.26_{\pm 0.44} \left ( \downarrow_{0.60} \right )$ \\
\textrm{w/ uni}         & $72.33_{\pm 1.55}$ & $80.24_{\pm 0.19}$ & $79.09_{\pm 0.35}$ & $73.02_{\pm 0.08}$ & $74.47_{\pm 0.16}$ & $73.62_{\pm 0.09}$ & $82.88_{\pm 0.08}$ & $76.52_{\pm 0.36} \left ( \downarrow_{2.34} \right )$ \\
\hline
\textrm{w/ cons}        & $80.39_{\pm 1.48}$ & $79.13_{\pm 0.25}$ & $78.86_{\pm 0.34}$ & $71.25_{\pm 0.22}$ & $72.08_{\pm 0.20}$ & $72.24_{\pm 0.22}$ & $84.14_{\pm 0.21}$ & $76.87_{\pm 0.42} \left ( \downarrow_{1.99} \right )$ \\
\textrm{w/ stab}         & $67.60_{\pm 1.54}$ & $70.35_{\pm 0.35}$ & $67.75_{\pm 0.16}$ & $69.99_{\pm 0.10}$ & $69.09_{\pm 0.14}$ & $68.46_{\pm 0.16}$ & $68.47_{\pm 0.19}$ & $68.82_{\pm 0.38} \left ( \downarrow_{10.04} \right )$ \\
\hline
\rowcolor[HTML]{E9E9E9}
\textbf{Full}   
& $\textbf{82.29}_{\pm 1.25}$ & $\textbf{81.78}_{\pm 0.27}$ & $\textbf{79.69}_{\pm 0.31}$ & $\textbf{74.26}_{\pm 0.17}$ & $74.25_{\pm 0.14}$ & $\textbf{75.08}_{\pm 0.21}$ & $\textbf{84.69}_{\pm 0.20}$ & $\textbf{78.86}_{\pm 0.36}$ \\
\bottomrule
\end{tabular}}
\label{tab:abl}
\end{table*}

\begin{figure}
\centering
\setlength{\belowcaptionskip}{0.2cm}
\subfloat[Trade-off parameter $\gamma$]{\label{fig:param1}\includegraphics[width=0.515\linewidth]{
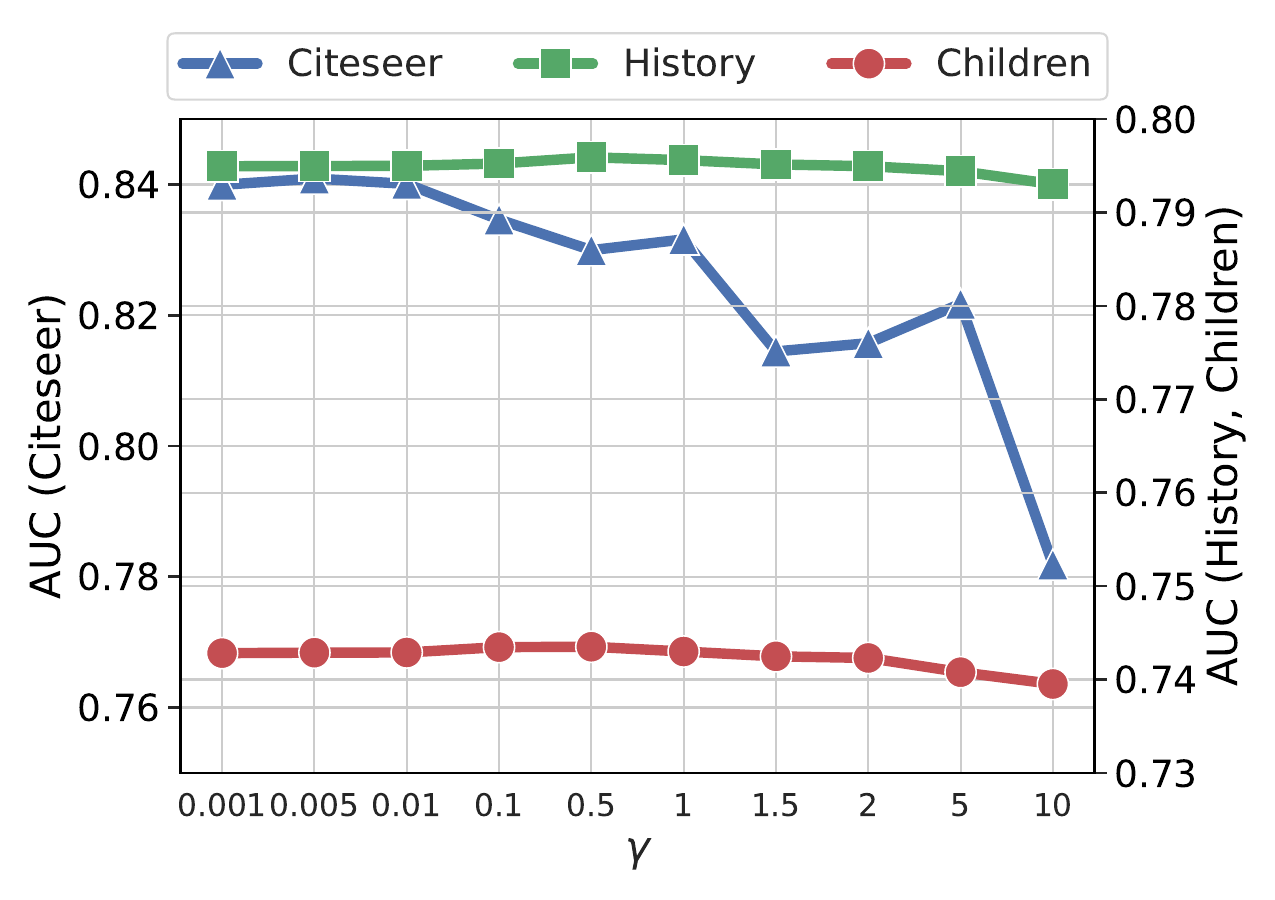}}
\subfloat[Sampling rounds $R$]{\label{fig:param2}\includegraphics[width=0.485\linewidth]{
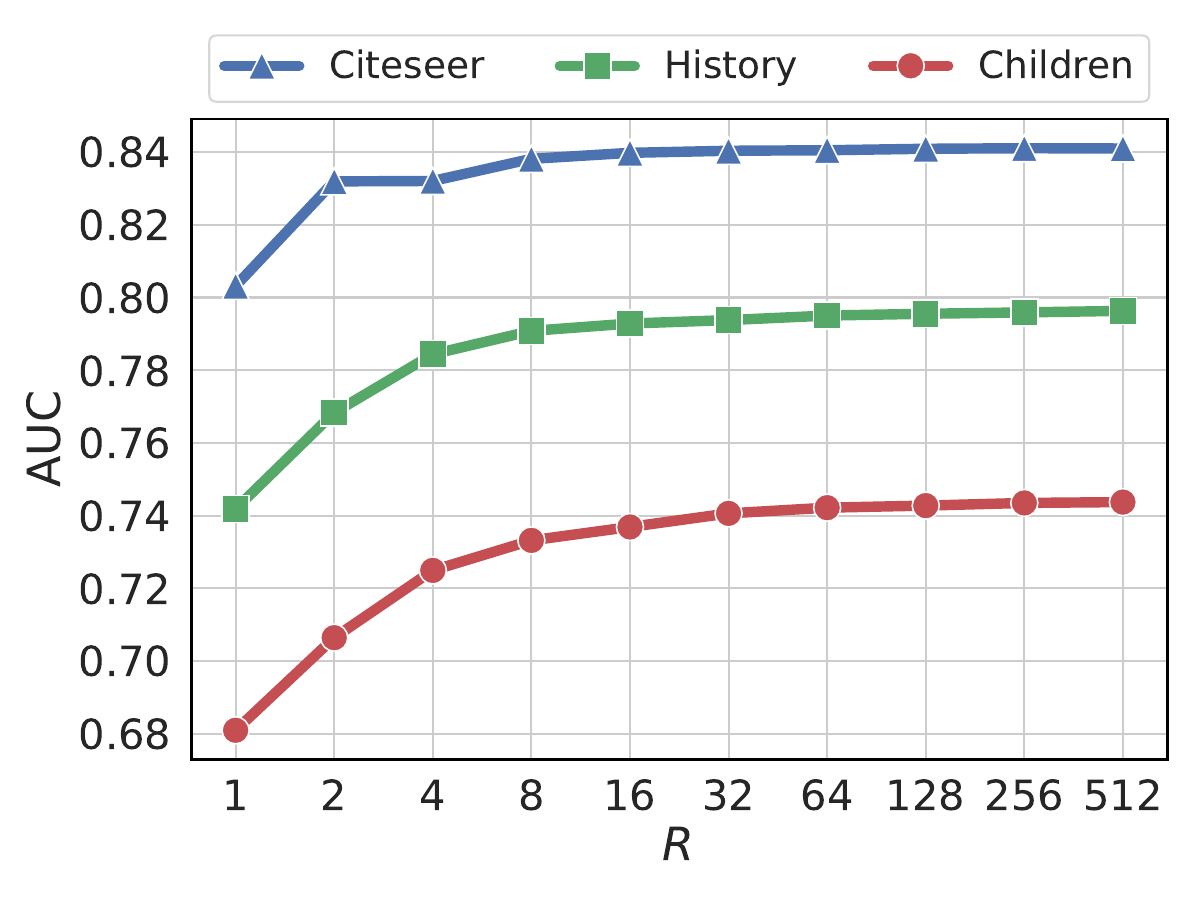}}
\caption{Sensitivity analysis for the trade-off parameter $\gamma$ and number of sampling rounds $R$ w.r.t. AUC.}

\label{fig:param}
\end{figure}

\begin{figure*}[ht]
\centering
\subfloat[Citeseer]{\label{fig:vis1}\includegraphics[width=.25\linewidth]{
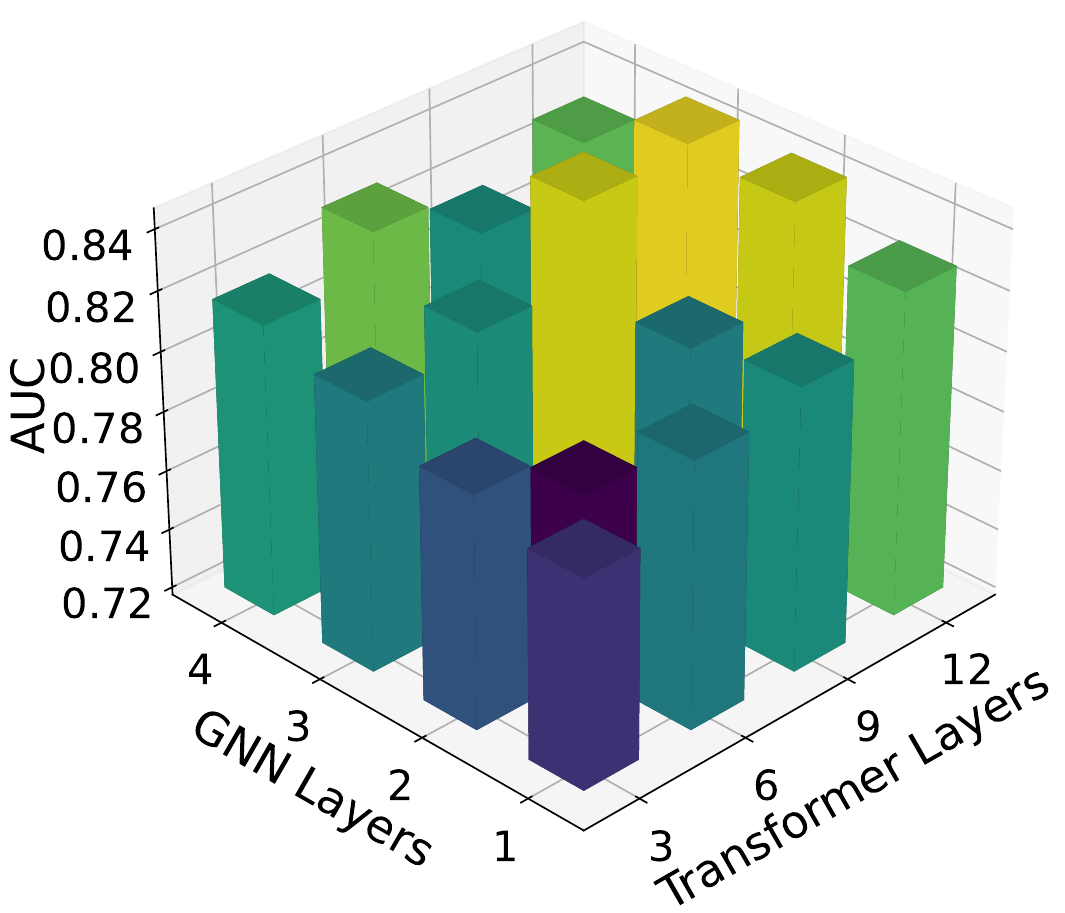}}
\subfloat[Pubmed]{\label{fig:vis2}\includegraphics[width=.25\linewidth]{
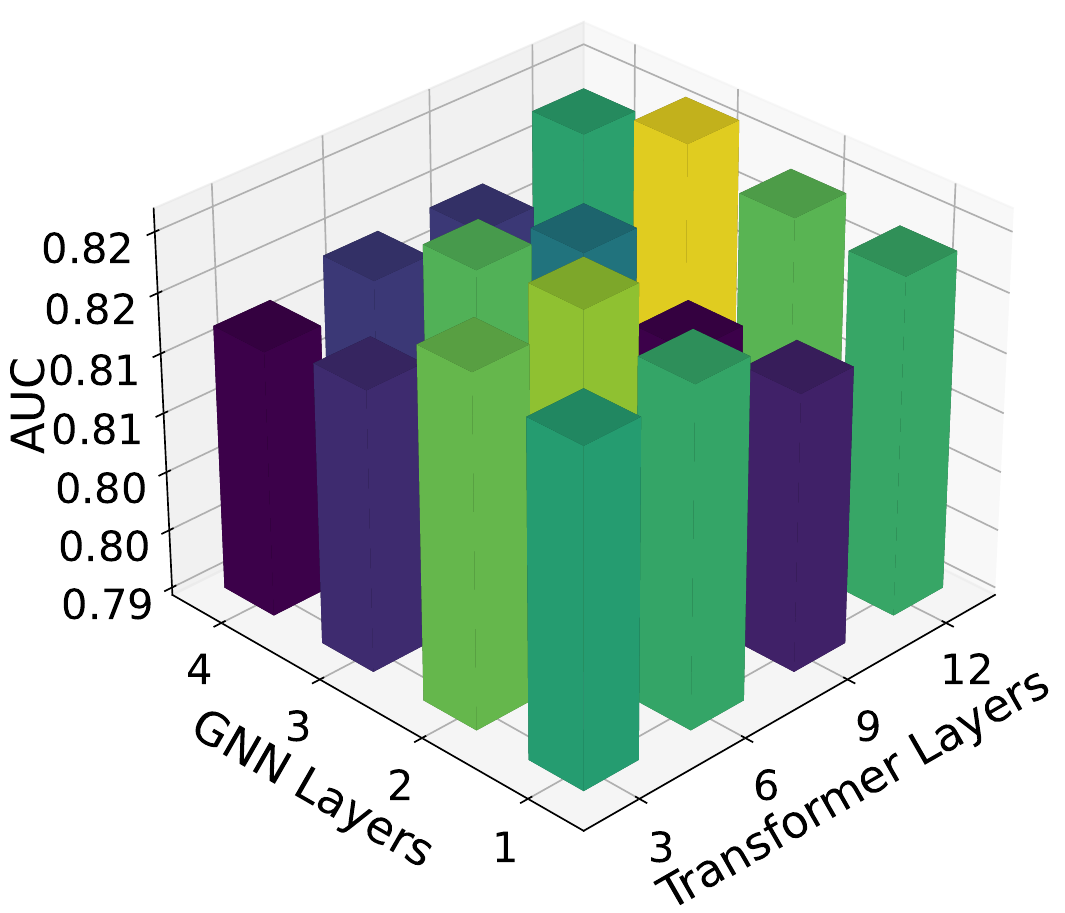}}
\subfloat[Computers]{\label{fig:vis4}\includegraphics[width=.25\linewidth]{
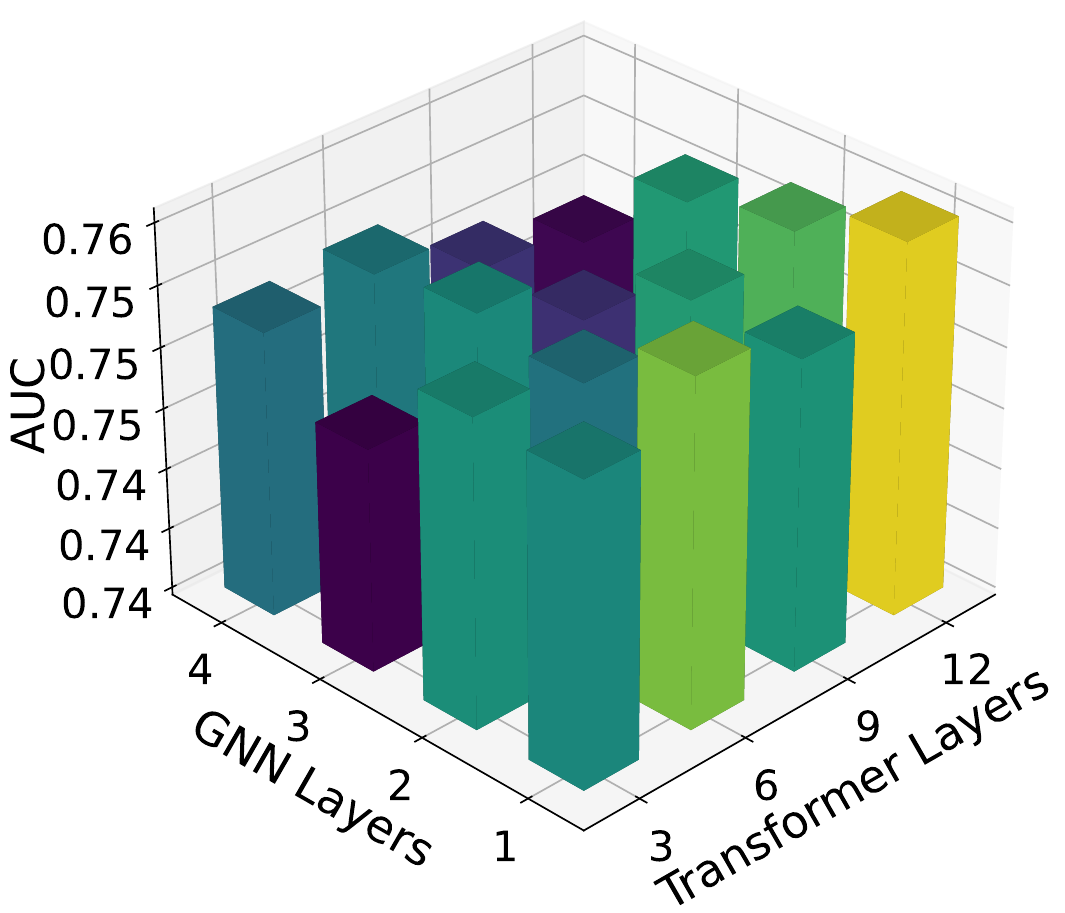}}
\subfloat[ogbn-Arxiv]{\label{fig:vis5}\includegraphics[width=.25\linewidth]{
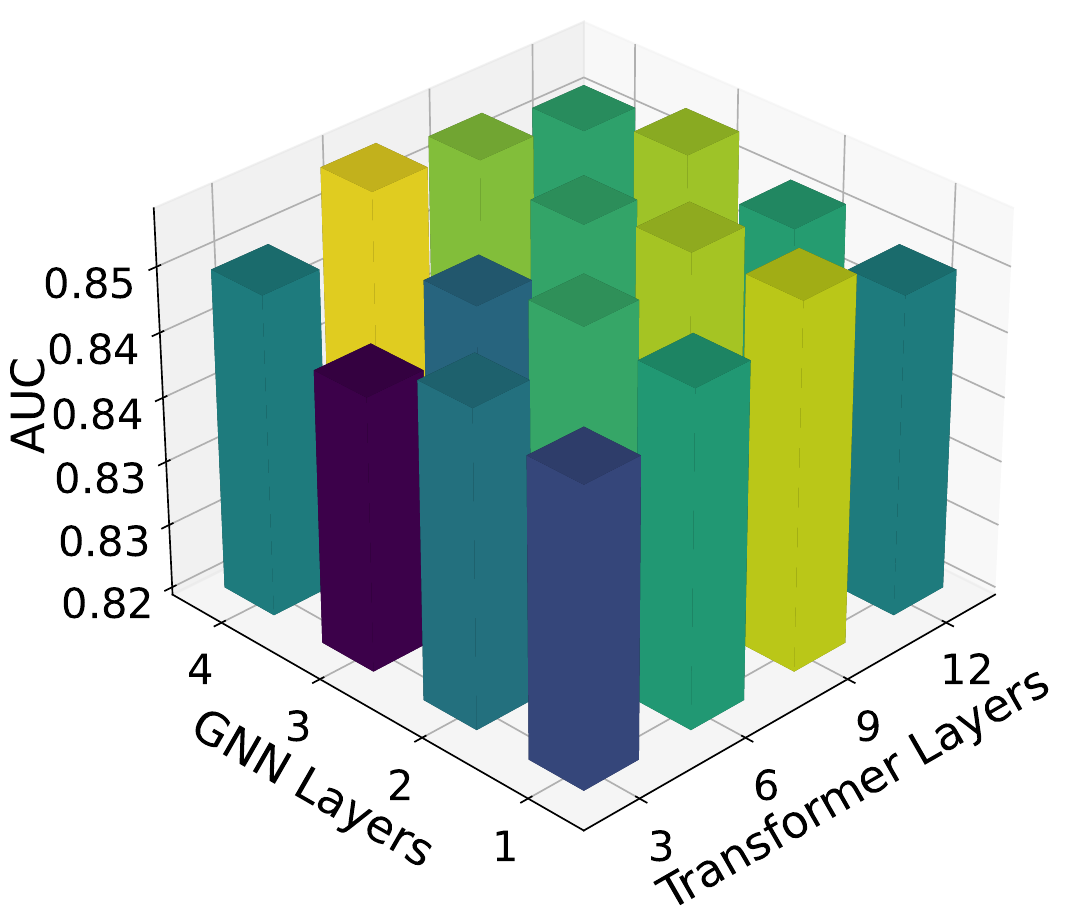}}\\	
\caption{Impact of layer configurations in bi-modal encoders w.r.t. AUC.}
\label{fig:param_layers}
\end{figure*}

\begin{figure*}[ht]
  \centering
  \includegraphics[width=1\linewidth]{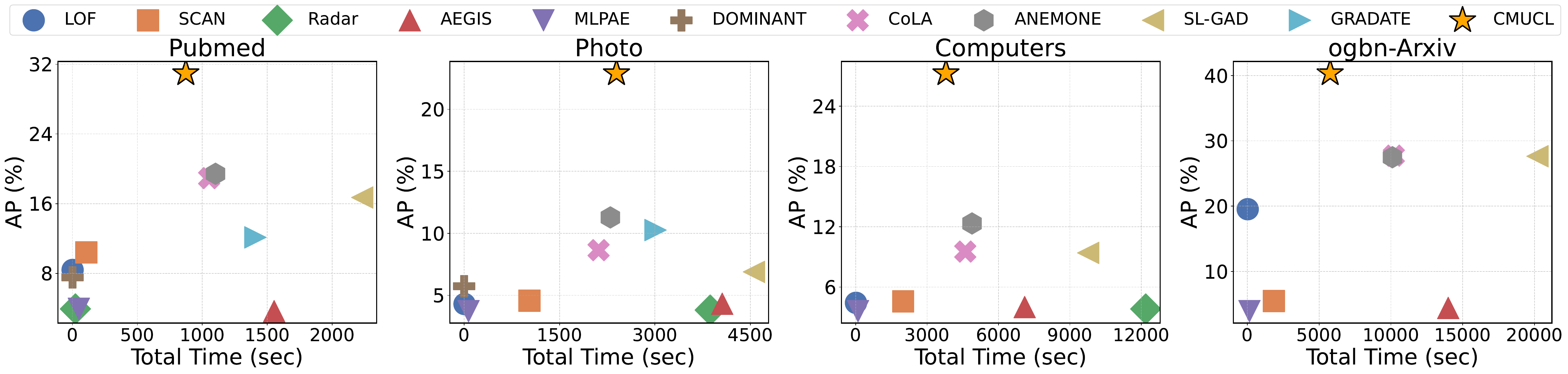}
  \caption{AP vs. Total Time (sec) for various methods across four datasets.}
  \label{fig:runtime}
\end{figure*}

\subsection{Result and Analysis}
We conduct an extensive anomaly detection study on eight datasets, comparing our method against eleven well-known approaches in a fully unsupervised setting. We employ ROC-AUC and average precision (AP) as evaluation metrics. As shown in Table~\ref{tab:ad}, our method outperforms the baseline models. Here are some key observations:

First, classical methods, such as density-based, clustering-based, and matrix factorization-based models, perform poorly overall. 
This underscores the limitations of non-deep learning approaches in capturing the complex feature and structural anomalies in text-attributed graphs. Deep learning-based graph methods are more competitive. However, reconstruction-based methods like DOMINANT, which require inputting the entire graph structure, fail to handle large datasets with 80k nodes and 720k edges, resulting in out-of-memory (OOM) errors. Contrastive learning-based methods, by designing various levels of contrast, effectively mine anomalous information and achieve state-of-the-art (SOTA) performance. 

However, existing works often overlook the supervision signals provided by textual features and utilize shallow methods to extract features that are inherently unrelated to anomalies, thereby falling into a suboptimal trap. By leveraging both cross-modal and uni-modal contrastive learning, CMUCL achieves the best performance in eight datasets, outperforming the runner-up by 4.68\% in AUC and 11.13\% in AP on average. Figure~\ref{fig:main_roc} further illustrates the effectiveness of our method in achieving high true positive rates and low false positive rates. Comprehensive results across all eight datasets are available in the supplementary material.
Furthermore, most reconstruction or contrastive learning-based methods cannot be applied to ultra-large-scale graph datasets, which hinders their application in real-world scenarios. Our method, however, does not require reading the entire graph structure and significantly improves performance over the second-best, demonstrating scalability and effectiveness on large-scale datasets. These results confirm that our method sets a new paradigm for anomaly detection performance on large-scale text-attributed graphs.

\subsection{Ablation Study}
\paragraph{Contrastive Strategy.}
We perform ablation studies to validate the effectiveness of the proposed cross-modal and uni-modal multi-scale contrastive approaches. For clarity, we denote the experiments using only cross-modal inner-scale contrast as \textrm{w/ cross inner}, those using only cross-modal inter-scale contrast as \textrm{w/ cross inter}, and those using uni-modal multi-scale contrast as \textrm{w/ uni}. As illustrated in Table~\ref{tab:abl}, it is evident that both \textrm{w/ cross inter} and \textrm{w/ uni} achieved superior performance, underscoring the critical importance of multi-scale approaches for anomaly detection. Our full model achieves the best performance, validating its effectiveness.

\paragraph{Anomaly Estimator Strategy.}
We further investigate the effects of consistency and stability in a train-free anomaly score estimator. The variant that considers only consistency in the estimator is \textrm{w/ cons} (Eq.~\eqref{eq:cons}). \textrm{w/ stab} refers to cases where only the instability of nodes across multiple detection rounds is considered, i.e., $\sqrt{\frac{\sum_{r=1}^{R} \left(s_i^r - \overline{S}_i \right)^{2}}{R}}$ in Eq.~\eqref{eq:estimator}. Table~\ref{tab:abl} shows that \textrm{w/ cons} achieves better performance. However, incorporating stability provides a more comprehensive assessment of the anomaly scores of nodes.

\subsection{Parameter Study}
\paragraph{Trade-off parameter $\gamma$.}
We explore the impact of the trade-off parameters $\gamma$ on model performance. As shown in Figure~\ref{fig:param1}, the model's performance initially increases with a rise in $\gamma$. However, when $\gamma$ reaches higher values and the uni-modal loss becomes dominant, the performance degrades, especially on the Citeseer dataset. From the ablation study in Table~\ref{tab:abl}, it can be seen that cross-modal inter-scale contrast is the most effective on Citeseer. Due to the effectiveness of cross-modal, $\gamma$ can usually be set to a smaller value. \looseness=-1

\paragraph{Sampling rounds $R$.}
We examine the impact of sampling rounds $R$ on the anomaly detection performance, as shown in Figure~\ref{fig:param2}. As $R$ increases, the performance increases significantly, which proves that a limited number of negative samples within a batch may lack sufficient discriminative power for effective anomaly detection. Beyond $R=64$, the incremental gains in performance plateau, indicating that further increases in $R$ yield diminishing returns. Consequently, we set $R$ to 256 across all datasets to optimize the balance between computational efficiency and detection efficacy.

\paragraph{Number of bi-model encoder layers.}
We investigate the impact of jointly trained text and graph encoders with varying layers on anomaly detection performance.  Our configurations include 1, 2, 3, and 4 layers of GCN encoders paired with 3, 6, 9, and 12 layers of transformers, as illustrated in Figure~\ref{fig:param_layers}. For the text encoders, we find that deeper Transformer layers, which enable more complex sequence transformations, generally significantly enhance the model’s ability to detect semantically relevant anomalies in the data. This result highlights the crucial role of a bi-encoder architecture in cross-modal learning for robust anomaly detection. For graph encoders, we observe that optimal performance is usually achieved with 2 or 3 layers. Therefore, deeper text encoders can be used in subsequent studies to mine anomalies in text-attributed graphs.

\subsection{Complexity Analysis}\label{sec:complexity_analysis}
In this study, we first analyze the time complexity of the proposed framework.
The time complexity is primarily attributable to the text encoder, the graph encoder, and the similarity calculation associated with the comparison pairs. The time complexity of the text encoder is $O(\left|\mathcal{V}\right|\overline{L}^{2}d)$, where $\left|\mathcal{V}\right|$ represents the number of nodes and $\overline{L}$ denotes the number of tokens in the text. The parameter $d$ represents the dimensionality of each token, which is equivalent to the output dimension of the graph encoder. The time complexity of the graph encoder is $O(\left|\mathcal{V}\right|d^2 + \left|\mathcal{E}\right|d)$, where $\left|\mathcal{E}\right|$ is the number of edges. The time complexity of the similarity calculation for all comparison pairs is $O(\left|\mathcal{V}\right|^2d)$.
In conclusion, the total time complexity of CMUCL is
$O\left(\left|\mathcal{V}\right|d(\left|\mathcal{V}\right| + \overline{L}^{2} + d) + \left|\mathcal{E}\right|d\right)$. It is typically the case that $\left|\mathcal{V}\right| \gg \overline{L}$. Therefore, the complexity is comparable to that of GAD methods based on contrastive learning $O\left(\left|\mathcal{V}\right|d(\left|\mathcal{V}\right| + d) + \left|\mathcal{E}\right|d\right)$.

We further present a quantitative analysis of runtime on eight benchmark datasets, as shown in Fig.~\ref{fig:runtime}. Our method achieves the best overall detection performance while maintaining efficient runtime, outperforming most deep learning baselines and only slightly slower than a few traditional methods. Detailed runtime measurements along with the AP results for the other four datasets and the AUC vs. total time are provided in the supplementary material.

Specifically, compared to GNN-based methods, traditional methods can typically only process a single type of data and have low model complexity, which results in faster processing time. However, due to their limited expressive capabilities, the detection results are often not ideal. For example, LOF and MLPAE can only process feature information, while SCAN solely handles structural information.
In contrast, although Radar considers both attributes and structure, it requires matrix inversion (\(O(\left|\mathcal{V}\right|^3)\)) when calculating the residual matrix, so its running time increases cubically with the increase of the dataset size.
In the context of GNN-based approaches, the reconstruction-based method DOMINANT exhibits excellent running time on small-scale graphs. Reconstruction-based methods typically require the input and reconstruction of the entire network structure, consuming significant memory. This results in rapid training on smaller graphs but limits its scalability on large-scale graph data and renders it inapplicable in practical application scenarios.
Existing contrastive methods (CoLA, ANEMONE, SL-GAD, and GRADATE) calculate consistency for each node with only one positive and one negative pair per epoch, leading to low efficiency. Our method employs a batch processing strategy during both the training and inference stages, where each contrastive view contains one positive and $N-1$ negative pairs, significantly improving the execution speed. 

Overall, CMUCL achieves optimal performance while striking an effective balance between accuracy and efficiency.

\section{Conclusion}
In this paper, we first explore anomaly detection in text-attributed graphs and construct eight datasets to advance future investigations. In addition, we propose a novel framework that integrates multi-scale cross- and uni-modal contrastive learning, moving beyond traditional methods to maximize leverage data potential for capturing subtle anomaly signals. Finally, we devise an anomaly score estimator that effectively evaluates node anomalies. Experiments validate the effectiveness of CMUCL in detecting anomalies in text-attributed graphs.



\begin{ack}
This research was partially supported by the Key Research and Development Project in Shaanxi Province No. 2024PT-ZCK-89, the National Science Foundation of China No. 62476215, 62302380, 62037001, 62137002 and 62192781, and the China Postdoctoral Science Foundation No. 2023M742789.
\end{ack}



\bibliography{mybibfile}

\clearpage
\begin{center}
    \huge \textbf{Appendix} \label{sec:appendix}
\end{center}
\appendix
\renewcommand\thefigure{A\arabic{figure}}
\renewcommand\theequation{A\arabic{equation}}
\frenchspacing
\maketitle

\section{Datasets details}\label{sec:appendix_datasets}
The eight widely used benchmark text-attributed graph datasets include four citation networks (Citeseer, Pubmed, ogbn-Arxiv, and CitationV8), and four e-commerce networks (History, Children, Photo, and Computers). The detailed descriptions of eight datasets are as follows:

\textbf{Citation Networks}. Citeseer, Pubmed~\cite{chen2024exploring}, ogbn-Arxiv, and CitationV8~\cite{yan2023comprehensive} are citation networks in which nodes represent academic papers and edges indicate citation information between these papers. The node attributes encompass the titles and abstracts of research papers.

\textbf{E-commerce Networks}. History, Children, Photo, and Computers~\cite{yan2023comprehensive} datasets are extracted from the Amazon dataset~\cite{ni2019justifying}. In these datasets, nodes represent various types of items, and edges signify items that are frequently purchased or browsed together. For the History and Children datasets, the node attributes are derived from the titles and descriptions of the respective books. Meanwhile, for the Photo and Computers datasets, the node attributes are sourced from high-rated reviews and product summaries.

To address the lack of explicitly labeled anomalies in existing text-attributed graph datasets, we follow standard construction methods from prior research~\cite{ding2019deep,liu2021anomaly,duan2023graph,duan2023graph} to develop a tailored anomaly labeling system, specifically designed for text-attributed graph anomaly detection, and apply it to adjust publicly available datasets.
Considering the datasets contain original textual content, we develop a novel method to generate both contextual and structural anomalies. The total number of anomalies for each dataset is presented in the final column of Table~\ref{tab:ad}. 

\textbf{Contextual anomaly.} Contextual anomalies refer to nodes whose attributes are demonstrably disparate from those of their neighboring nodes~\cite{song2007conditional,ma2021comprehensive}. We design two novel strategies, insertion and replacement, to perturb the original textual attributes of nodes to generate contextual anomalies. 
To generate such anomalies, we first randomly select a target node $v_i$ and then sample a set of $K$ nodes as the candidate set. We employ the BGE~\cite{bge_embedding} to encode textual information into attribute vectors and calculate the cosine similarity between $v_i$ and each node in the candidate set. Subsequently, we select the node $v_j$ with the lowest similarity in the candidate set as the source of abnormal information. The first strategy is to insert a specified segment of text from $v_j$ into a randomly selected position within the text of $v_i$. Another approach is to randomly replace the text. This entails randomly selecting an equal number of sentences from $v_i$ and $v_j$, and replacing the corresponding sentences from $v_i$ with those from $v_j$. Both insertion and replacement strategies construct the same number of contextual anomalies. Here, we set $K = 50$ to ensure the disturbance amplitude is large enough.

\begin{figure*}
\centering
\subfloat[Citeseer]{\label{fig:Degree_Citeseer_inj}\includegraphics[width=.25\linewidth]{
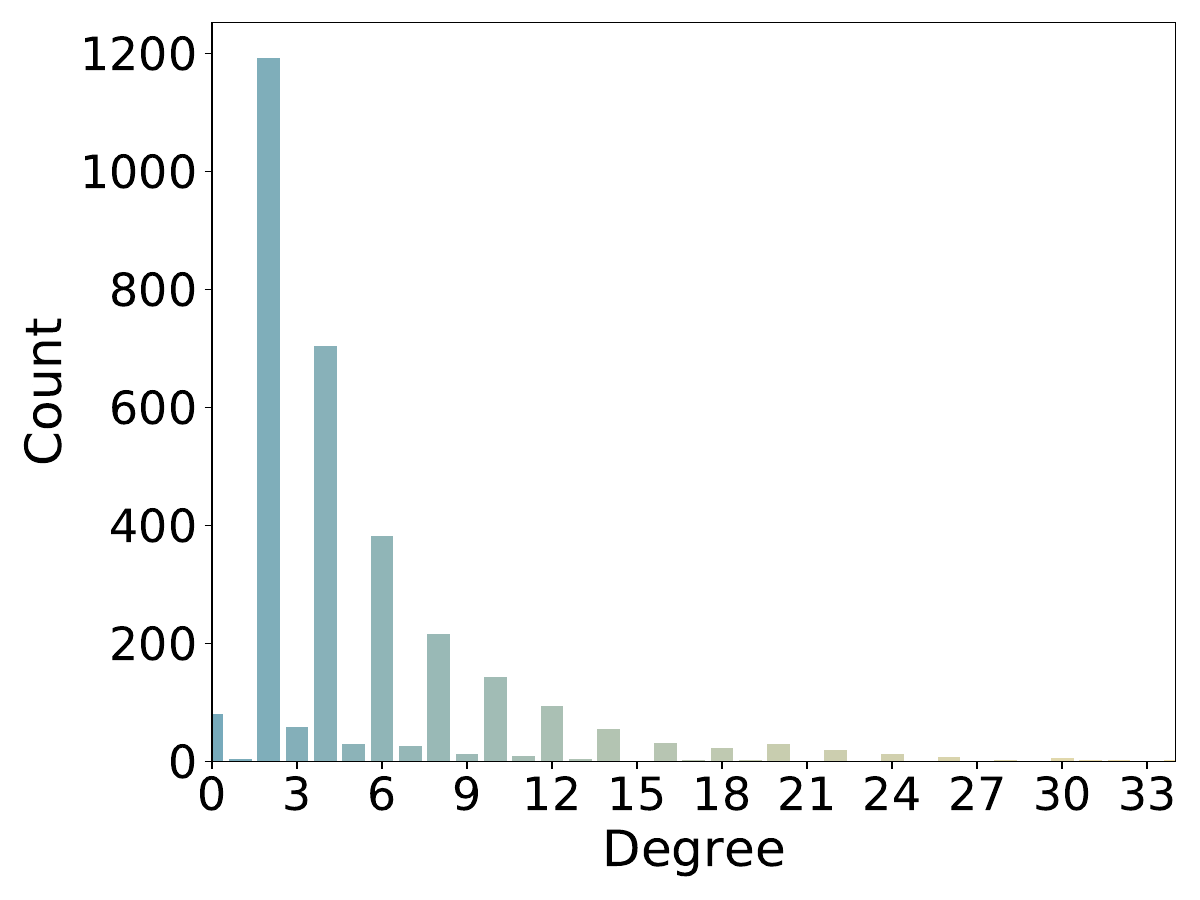}}
\subfloat[Pubmed]{\label{fig:Degree_Pubmed_inj}\includegraphics[width=.25\linewidth]{
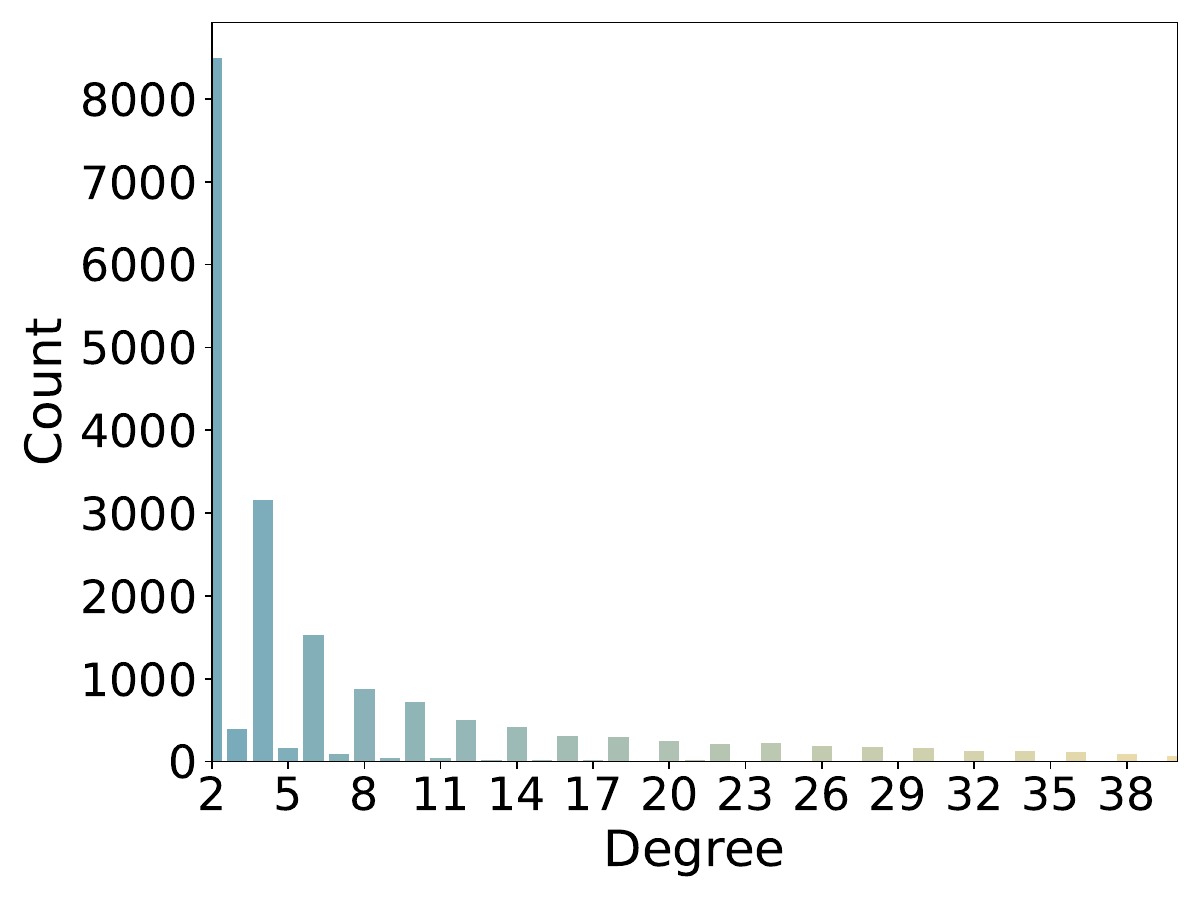}}
\subfloat[History]{\label{fig:Degree_History_inj}\includegraphics[width=.25\linewidth]{
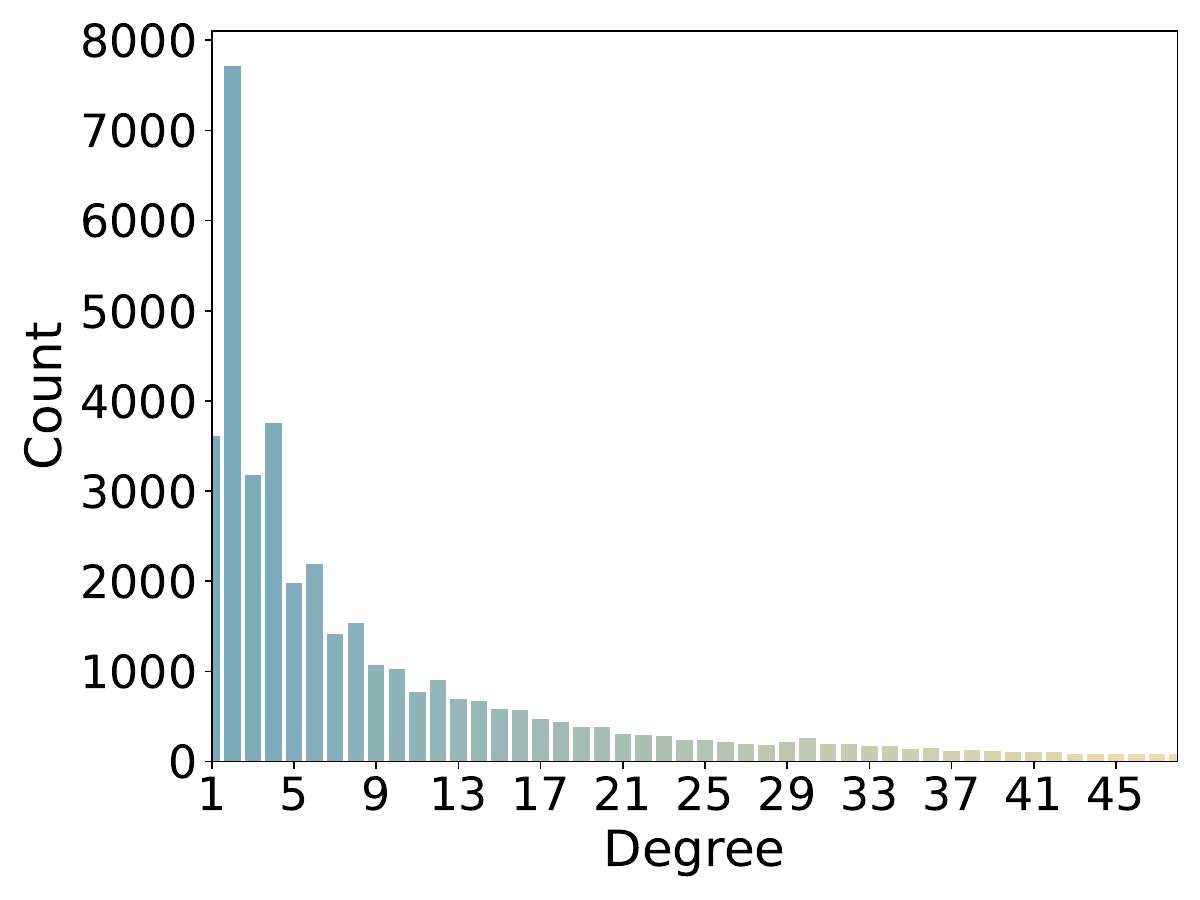}}	
\subfloat[Photo]{\label{fig:Degree_Photo_inj}\includegraphics[width=.25\linewidth]{
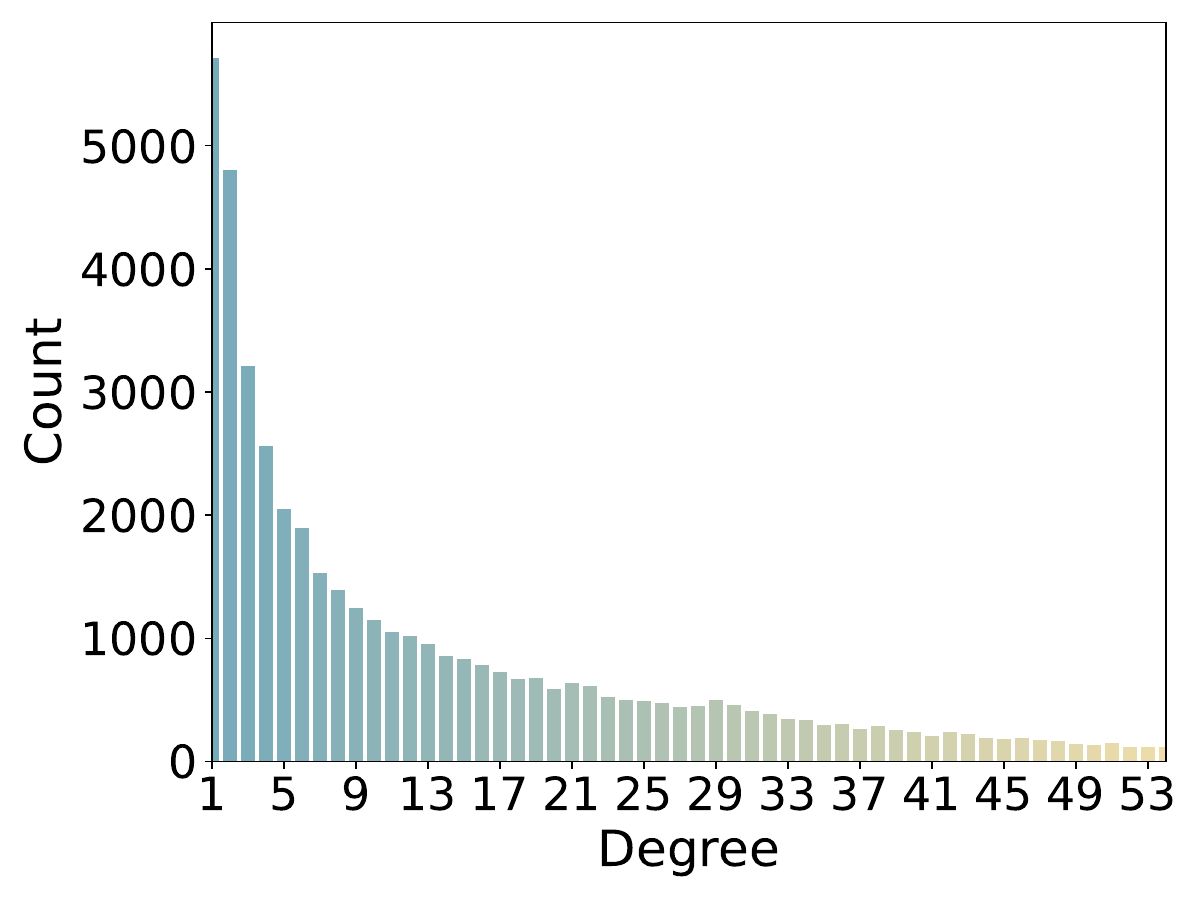}}\\	
\subfloat[Computers]{\label{fig:Degree_Computers_inj}\includegraphics[width=.25\linewidth]{
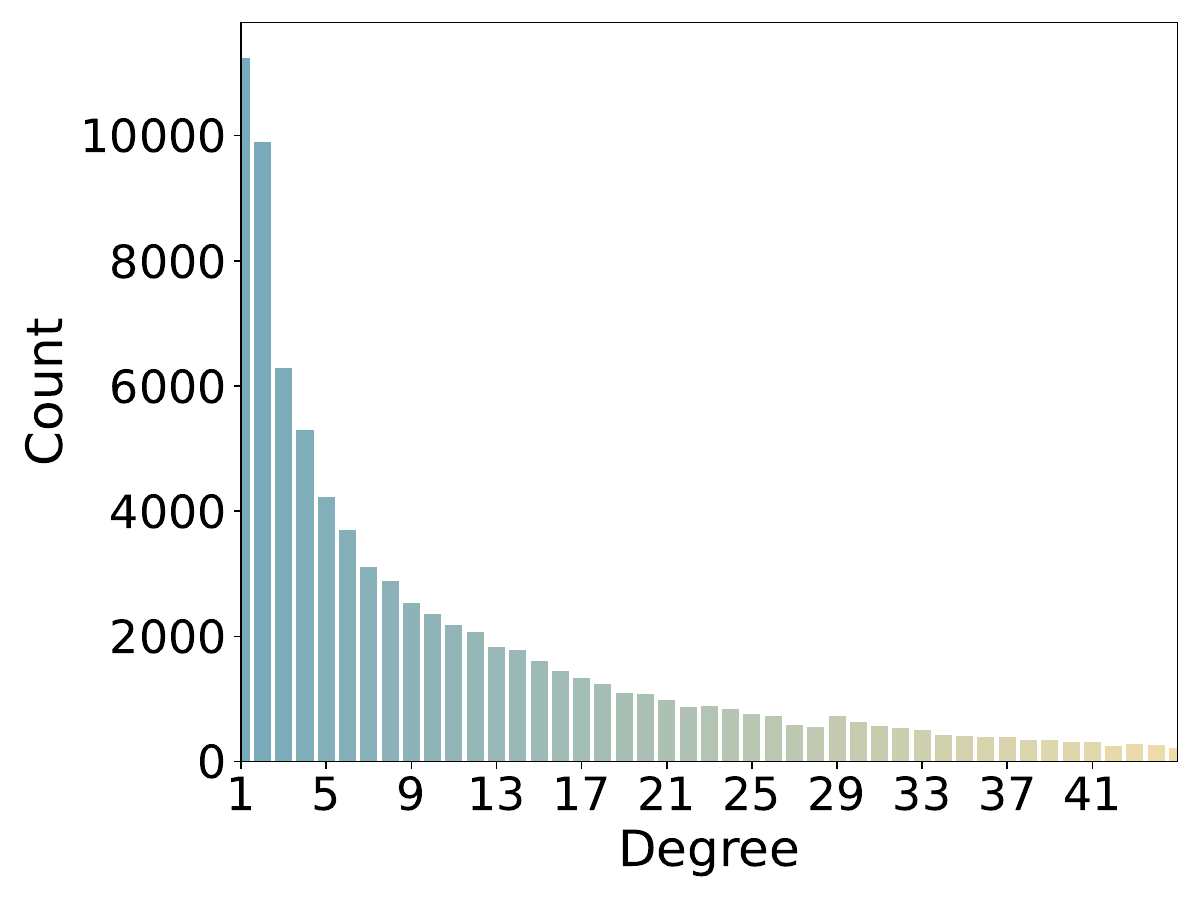}}
\subfloat[Children]{\label{fig:Degree_Children_inj}\includegraphics[width=.25\linewidth]{
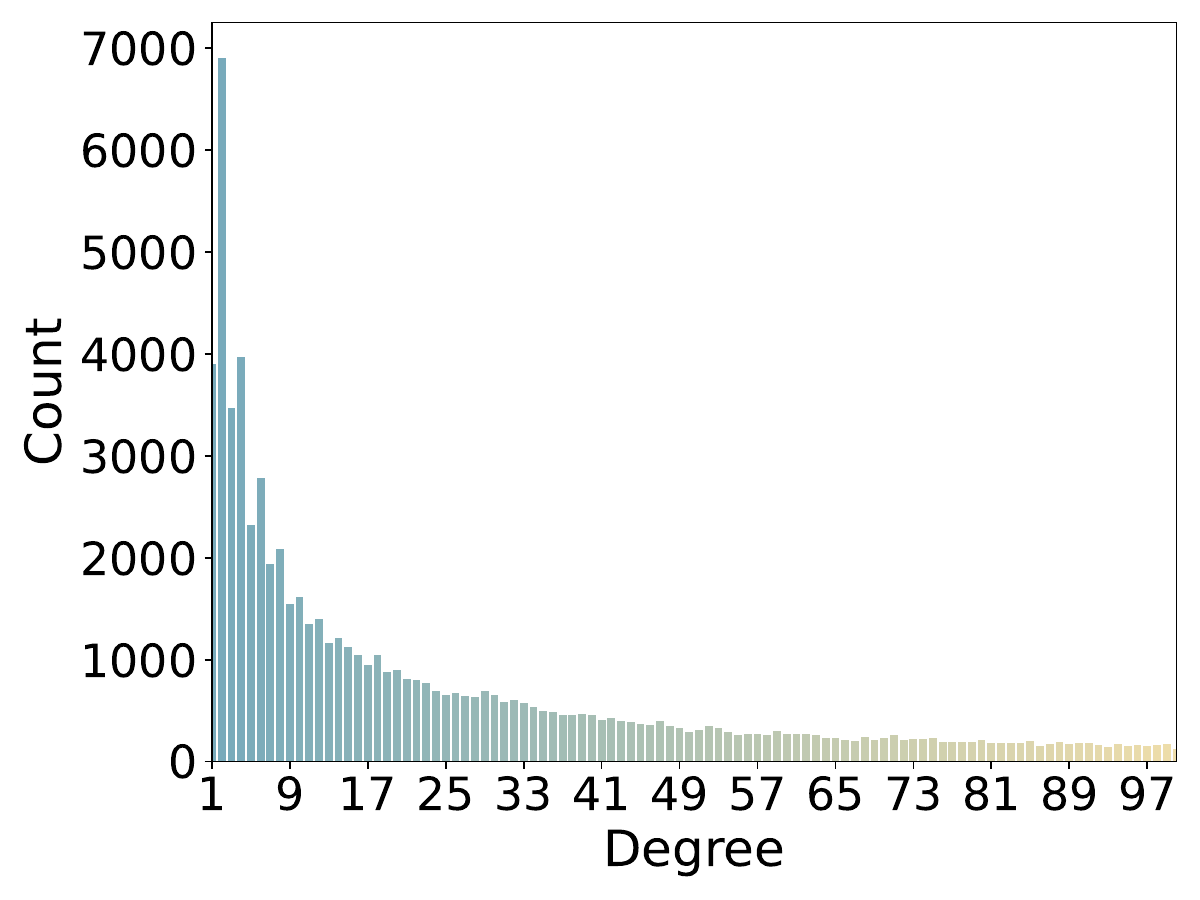}}
\subfloat[ogbn-Arxiv]{\label{fig:Degree_ogbn-Arxiv_inj}\includegraphics[width=.25\linewidth]{
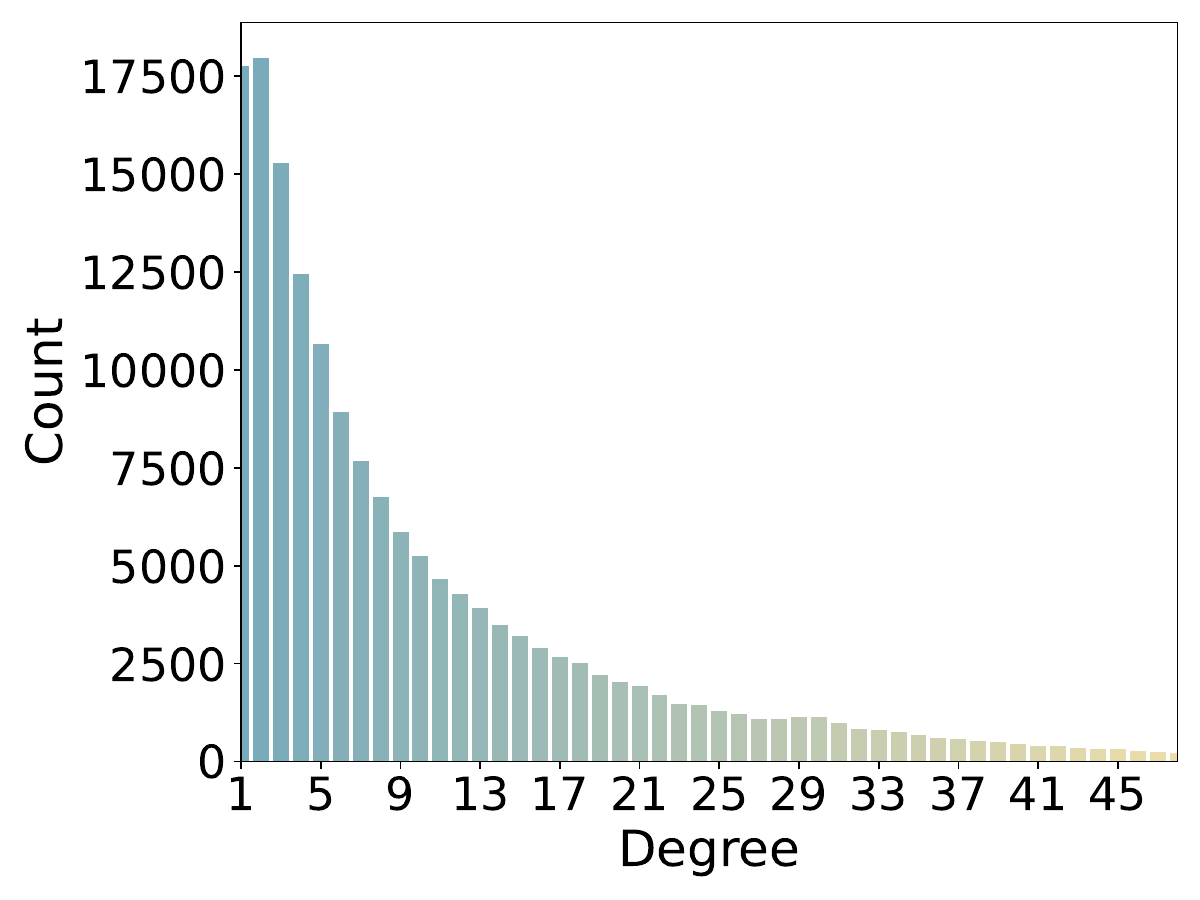}}
\subfloat[CitationV8]{\label{fig:Degree_CitationV8_inj}\includegraphics[width=.25\linewidth]{
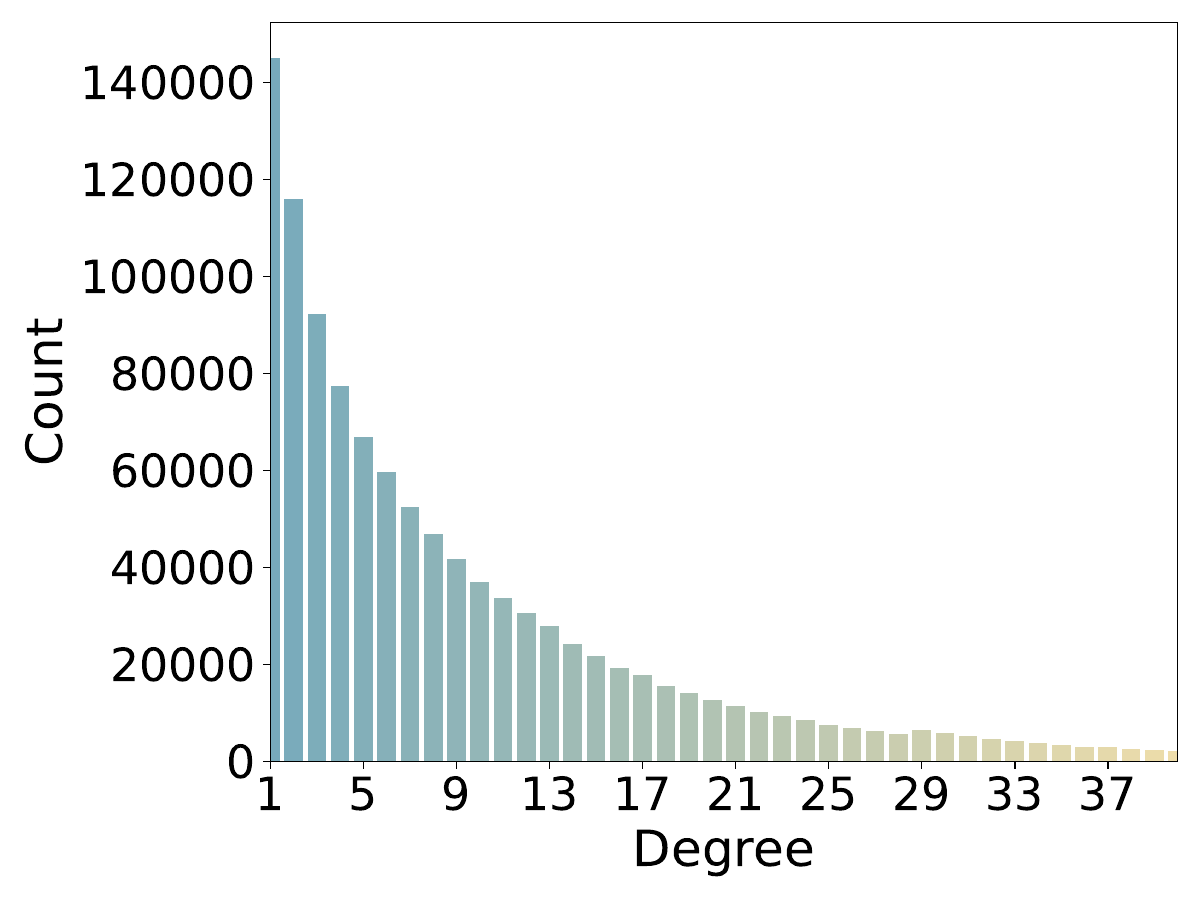}}\\
\caption{Degree distributions for eight benchmark datasets.}
\label{fig:degree_distributions}
\end{figure*}

\textbf{Structural anomaly.} Structural anomaly nodes usually have different connection patterns~\cite{ma2021comprehensive}, such as forming dense connections with others or connecting different communities. Therefore, we also design two strategies in this study to model these two types of structural anomalies. In real-world networks, a typical structural anomaly occurs when connections among nodes within a small clique are significantly denser than average~\cite{skillicorn2007detecting}. Thus, the first strategy injects structural anomalies that form dense connections with others. The process begins with the random selection of $q$ nodes and fully connecting them to form a clique. This step is repeated $p$ times to create $p$ such cliques, each consisting of $q$ nodes. In addition, anomaly nodes often build relationships with many benign nodes to boost their reputation and gain undue benefits, a behavior seldom seen among benign nodes~\cite{pandit2007netprobe,shin2017densealert}. Therefore, the second strategy injects structural anomalies that connect different communities by randomly adding edges. 
We start by randomly selecting a target node $v_i$. We then randomly add different numbers of edges to $v_i$ to generate structural anomalies that connect different communities. The number of edges for each target node $v_i$ is determined by sampling from the degree distribution of the original graph dataset. This approach ensures that the newly added structural anomalies continue to exhibit statistical characteristics aligned with those of the original graph, such as the degree distributions shown in Figure~\ref{fig:degree_distributions}. Assuming the total number of abnormal nodes is $4m$, $m$ abnormal nodes are injected for each of the aforementioned strategies.

\section{Implementation Details} \label{sec:appendix_param}
We report the mean and standard deviation of the results of all experiments run 5 times using different randomized seeds.
The computing infrastructure used for running experiments: Ubuntu 22.04, CPU: AMD EPYC 7542 32-Core, GPU: NVIDIA 4090, CUDA: 12.2, Memory: 500Gi.
Due to the considerable size of CitationV8, we employs a GPU with large memory when processing this dataset. The particular execution environment is as follows: Ubuntu 20.04, CPU: Intel(R) Xeon(R) Platinum 8163. GPU: CUDA 11.6, NVIDIA A100, Memory: 500Gi. 
In addition, versions of relevant software libraries and frameworks: Python: 3.8.13, torch: 1.12.1, torch-cluster: 1.6.0, torch-geometric: 2.1.0.post1, torch-scatter: 2.0.9, torch-sparse: 0.6.15, torch-spline-conv: 1.2.1, torchaudio: 0.12.1, torchvision: 0.13.1, transformers: 4.24.0, DGL: 0.9.0.
Finally, the Range of values tried per parameter during development: the learning rate parameter is selected from \{1e-5, 2e-5, 5e-5, 2e-4\}, epoch is selected from \{1, 2, 3\}, trade-off parameter $\gamma$ is selected from \{0.001, 0.005, 0.01, 0.5, 1.0, 1.5, 2.0, 5.0, 10.0 \}, and sampling rounds $R$ is select from \{1, 2, 4, 8, 16, 32, 64, 128, 256, 512\}. 
Specifically, the learning rates, $\gamma$ values, and the number of epochs as follows: Citeseer (2e-4, 5e-3, 2), Pubmed (2e-5, 1e-3, 2), History (2e-5, 0.5, 2), Photo (5e-5, 1e-3, 3), Computers (2e-5, 1e-2, 3), Children (5e-5, 0.5, 2), ogbn-Arxiv (1e-5, 1e-2, 2), and CitationV8 (2e-5, 0.5, 2).

\begin{table}[ht]
\centering
\caption{Comparison of average AUC for contextual and structural anomaly detection performance. CMUCL achieves the best performance across both categories.}
\label{tab:mean_auc_comparison_updated}
\resizebox{0.5\textwidth}{!}{
\begin{tabular}{lcc}
\toprule
\textbf{Method} & \textbf{Contextual Anomalies} & \textbf{Structural Anomalies} \\
\midrule
DOMINANT  & 46.67 & 78.21 \\
GAD-NR    & 48.60 & 66.27 \\
CoLA      & 59.16 & 91.09 \\
ANEMONE   & 59.30 & 92.37 \\
GRADATE   & 54.00 & 86.06 \\
\rowcolor[HTML]{E9E9E9}
\textbf{CMUCL} & \textbf{65.09} & \textbf{94.66} \\
\bottomrule
\end{tabular}
}
\end{table}

\section{Supplementary Main Results}
\begin{figure*}[ht]
    \centering
    \subfloat[Citeseer]{\label{fig:roc1}\includegraphics[width=0.25\linewidth]{
    figures/roc/ROC_Citeseer.pdf}}
    \subfloat[Pubmed]{\label{fig:roc2}\includegraphics[width=0.25\linewidth]{
    figures/roc/ROC_Pubmed.pdf}}
    \subfloat[History]{\label{fig:roc3}\includegraphics[width=0.25\linewidth]{
    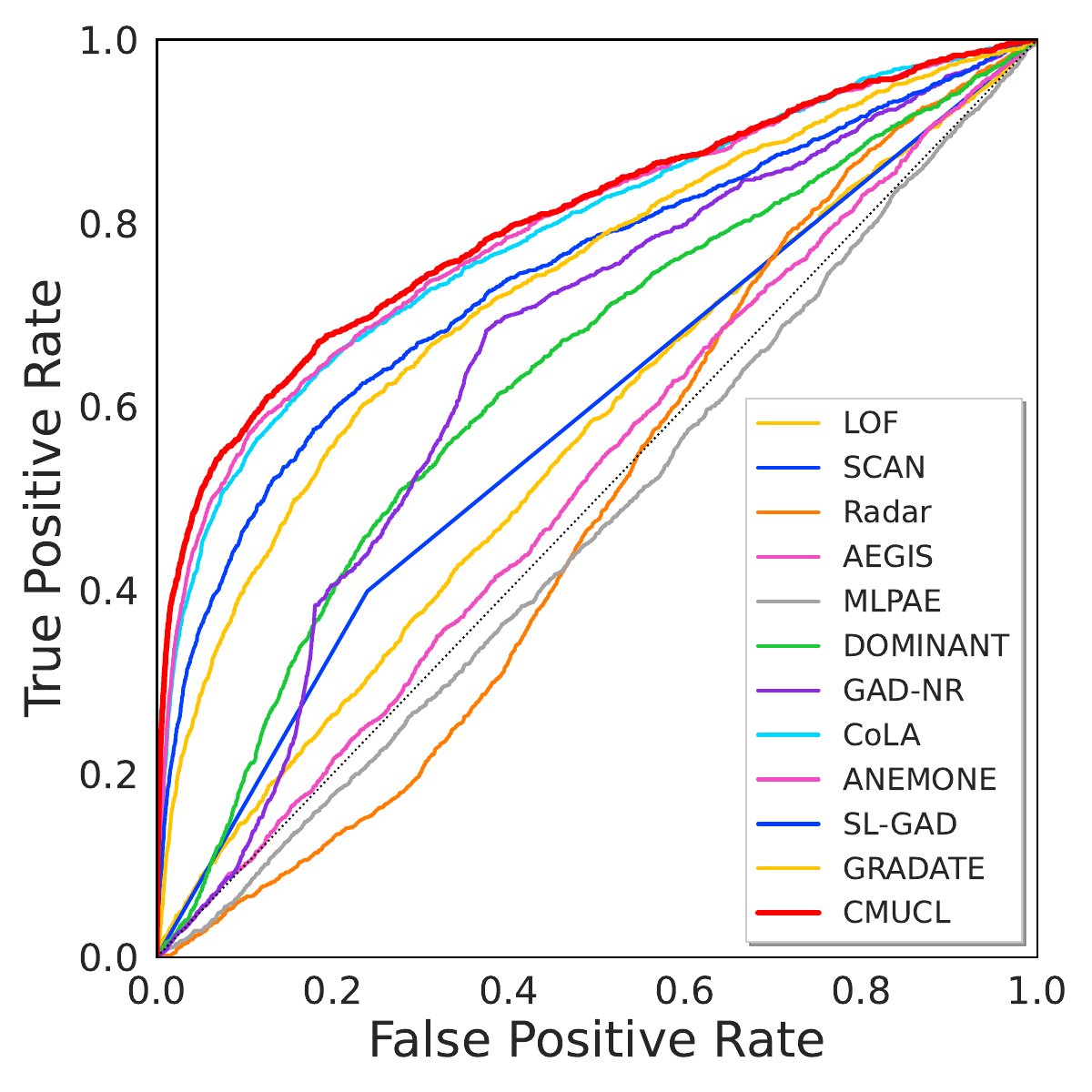}}
    \subfloat[Photo]{\label{fig:roc4}\includegraphics[width=0.25\linewidth]{
    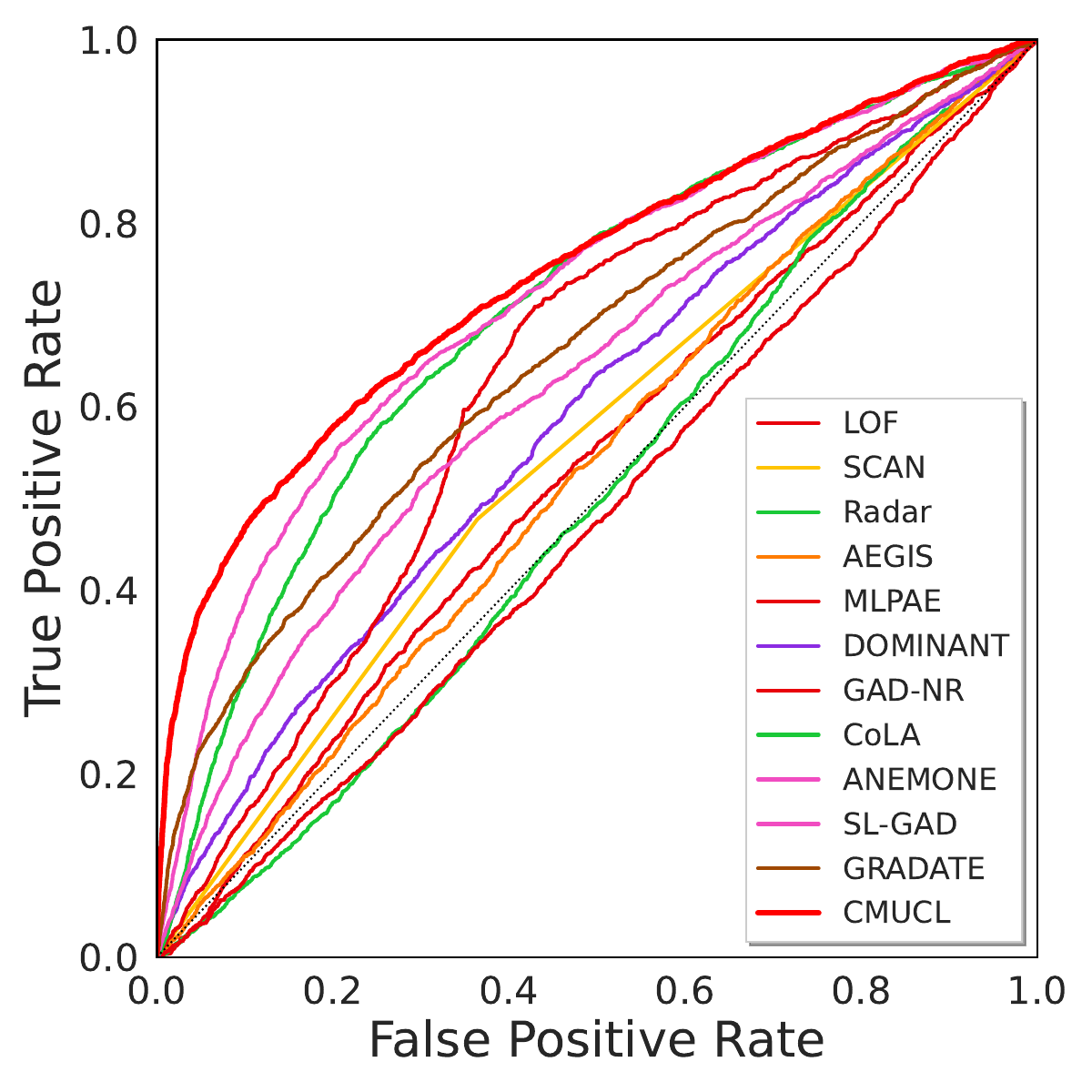}}\\
    \subfloat[Computers]{\label{fig:roc5}\includegraphics[width=0.25\linewidth]{
    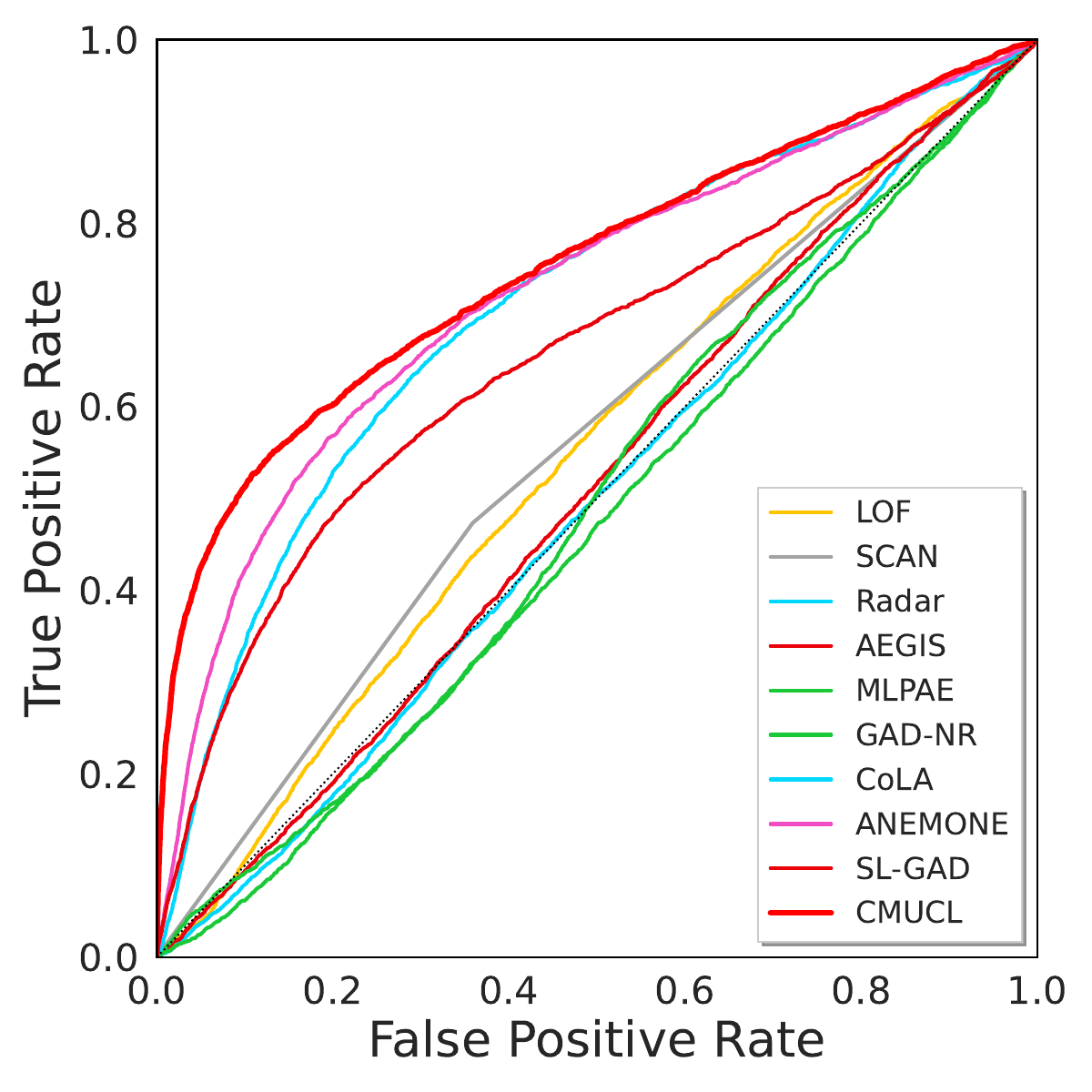}}
    \subfloat[Children]{\label{fig:roc6}\includegraphics[width=0.25\linewidth]{
    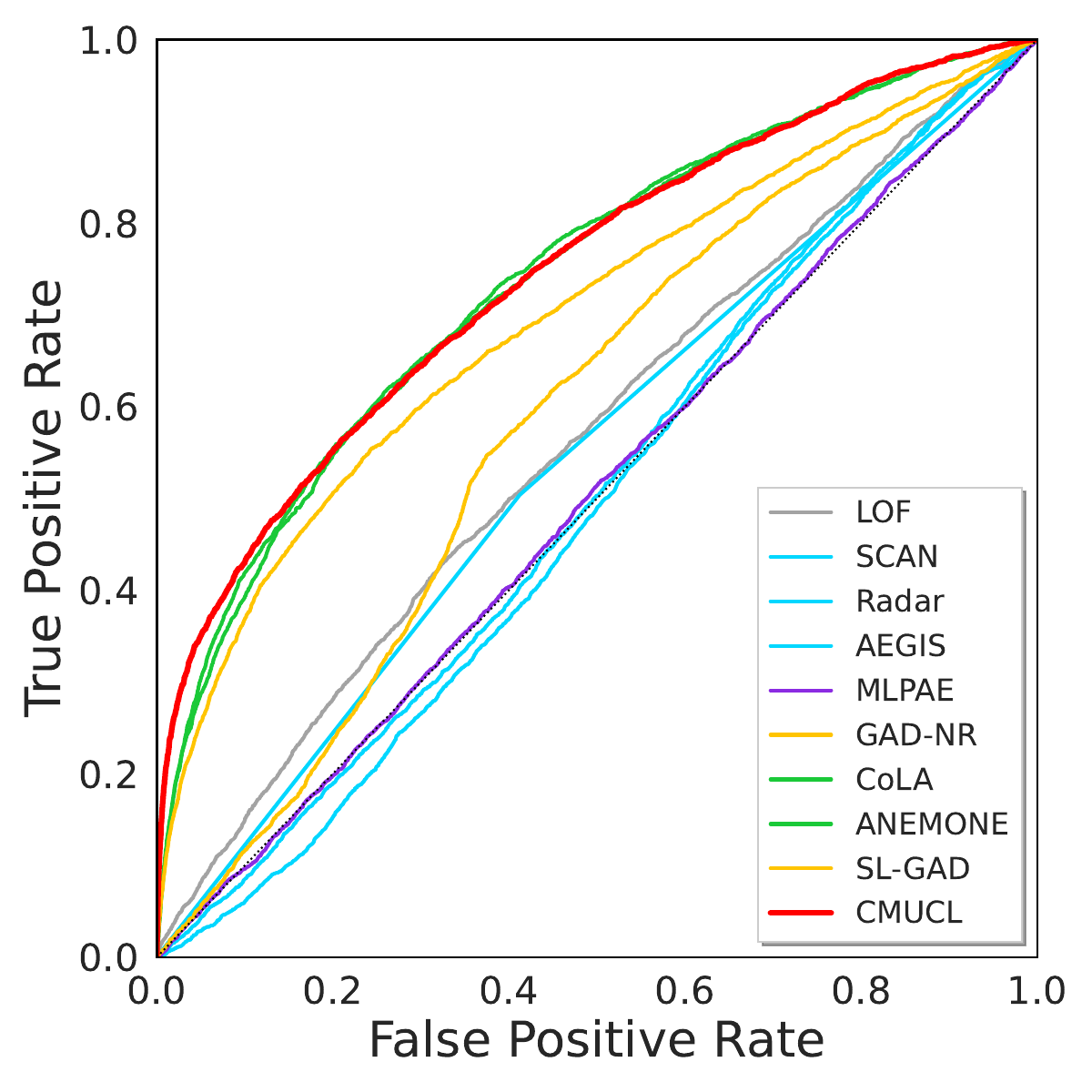}}
    \subfloat[ogbn-Arxiv]{\label{fig:roc7}\includegraphics[width=0.25\linewidth]{
    figures/roc/ROC_Arxiv.pdf}}
    \subfloat[CitationV8]{\label{fig:roc8}\includegraphics[width=0.25\linewidth]{
    figures/roc/ROC_CitationV8.pdf}}
    \caption{ROC curves compared on eight datasets. A larger area under the curve means better performance. The black dotted lines show the performance of random guessing.}
    \label{fig:roc}    
\end{figure*}

The experimental results presented in Figure~\ref{fig:roc} demonstrate the superior performance of our proposed method, across all eight datasets. CMUCL consistently achieves the highest ROC curve area, indicating its strong capability in detecting anomalies with high true positive rates and low false positive rates. Notably, in complex datasets such as ogbn-Arxiv and CitationV8, CMUCL achieves the most substantial improvements, highlighting its effectiveness in modeling large-scale and intricate graph structures. On traditional citation networks like Citeseer and Pubmed, CMUCL also outperforms other methods, showcasing its adaptability to sparse scenarios. \looseness=-1

We further analyze the performance of our method and existing approaches in detecting contextual and structural anomalies. We provide the average AUC scores for detecting these two types of anomalies across datasets, comparing CMUCL with five competing methods. During the computation of AUC for one anomaly type, nodes of the other type are masked in both predictions and labels. The results show that our method outperforms existing approaches in detecting both contextual and structural anomalies. These findings highlight the robustness and adaptability of CMUCL across different anomaly types. Interestingly, all methods perform better at capturing structural anomalies, suggesting that improving the detection of contextual anomalies is an important direction for future research.

In summary, CMUCL demonstrates remarkable generalizability and scalability, making it a leading approach for anomaly detection in text-attributed graphs. Its ability to integrate multimodal data effectively and capture subtle anomalies sets it apart from existing methods.

\begin{table}\small
\centering
\caption{AUC comparison between CMUCL and its variant without entropy minimization across four datasets.}
\label{tab:supp_abl}
\begin{tabular}{lcccc}
\toprule
\textbf{Method} & \textbf{Citeseer} & \textbf{Pubmed} & \textbf{History} & \textbf{Photo} \\ \midrule
w/o entropy     & 80.57             & 81.70           & 79.58            & 72.67          \\
\rowcolor[HTML]{E9E9E9}
CMUCL           & \textbf{82.29}    & \textbf{81.78}  & \textbf{79.69}   & \textbf{74.26} \\ \bottomrule
\end{tabular}
\end{table}

\section{Supplementary Ablation Study}
To further understand the contribution of entropy minimization in Eq.~\eqref{eq:ase}, we compare the full CMUCL model with its variant without entropy minimization in four data sets: Citeseer, Pubmed, History, and Photo.

As shown in Table~\ref{tab:supp_abl}, the ablation results indicate that removing entropy minimization in the anomaly score estimator consistently leads to a drop in AUC across four datasets. Specifically, the AUC drop is more pronounced on the Citeseer and Photo datasets, with a decrease of 1.72\% and 1.59\%, highlighting the importance of entropy minimization in enhancing detection performance in more challenging and complex scenarios. Similarly, smaller but consistent performance declines are observed on the Pubmed and History datasets.

Entropy minimization likely enhances the model's ability to better separate normal and anomalous nodes by encouraging confident predictions. These results validate the effectiveness of incorporating entropy minimization into the proposed framework, as it plays a crucial role in optimizing performance across diverse datasets.

\section{Complexity Analysis Results}\label{sec:appendix_complexity_analysis}
\begin{figure*}
  \centering
  \includegraphics[width=1\linewidth]{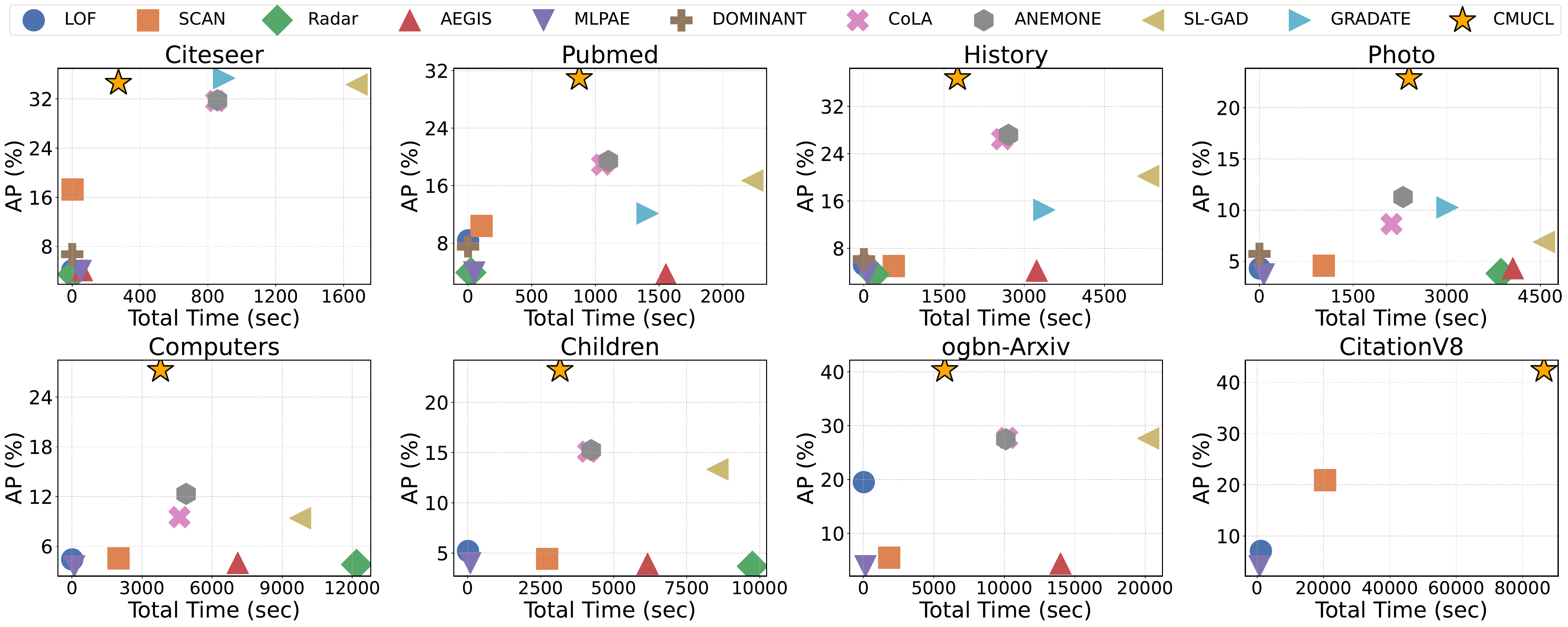}
  \caption{AP vs. Total Time (sec) for various methods across datasets.}
  \label{fig:appendix_runtime}
\end{figure*}

\begin{figure*}
  \centering
  \includegraphics[width=1\linewidth]{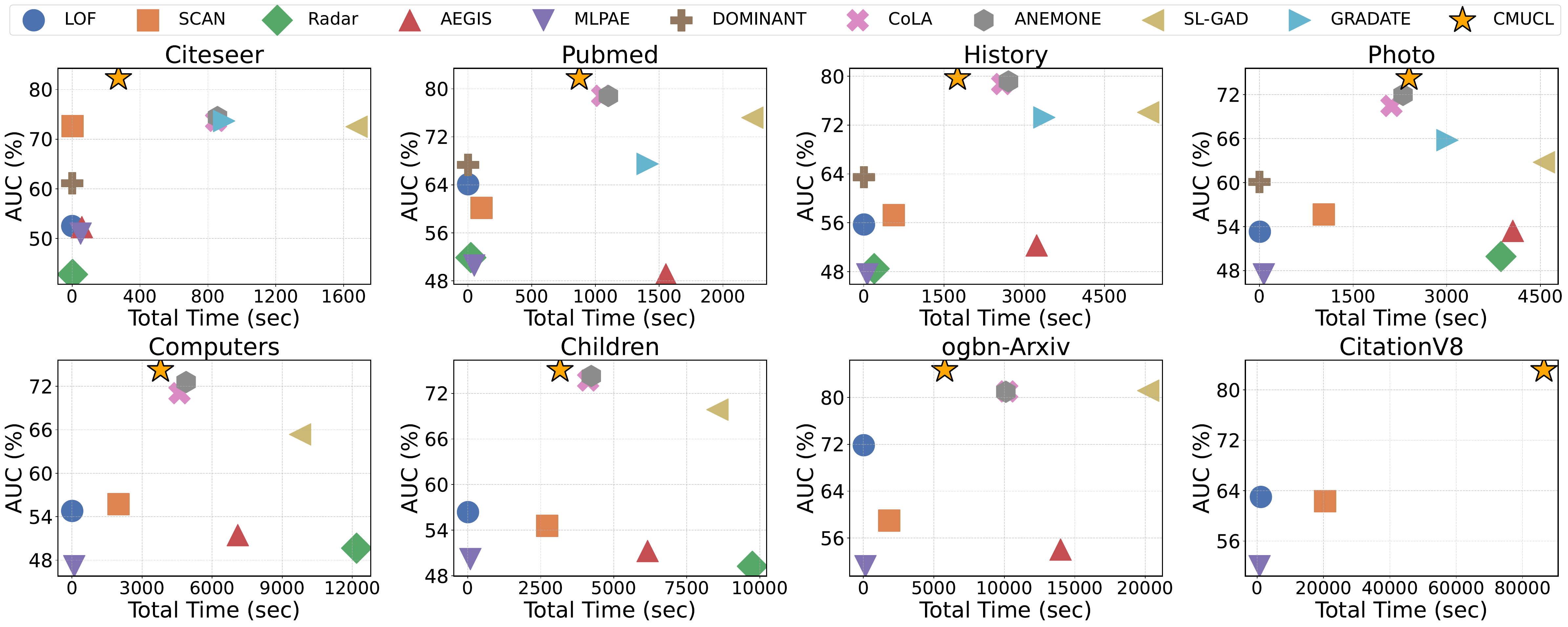}
  \caption{AUC vs. Total Time (sec) for various methods across datasets.}
  \label{fig:appendix_runtime1}
\end{figure*}

We show the results of AP and AUC vs. total time on 8 datasets in Figures~\ref{fig:appendix_runtime} and Figures~\ref{fig:appendix_runtime1} respectively. The specific running time is provided in Table~\ref{tab:time}.

\begin{table*}
\renewcommand\arraystretch{1.1}
\centering
  \caption{A comparative analysis of the running time for eight datasets. All measurements are in seconds.}
  \label{tab:time}
  \begin{tabular}{ccccccccc}
    \toprule
    Method & Citeseer & Pubmed & History & Photo & Computers & Children & ogbn-Arxiv & CitationV8 \\
    \midrule
LOF & 0.63 & 1.44 & 4.48 & 5.29 & 9.53 & 7.23 & 28.93 & 1,109.04  \\
SCAN & 2.65 & 107.52 & 561.90 & 1,027.66 & 1,991.91 & 2,717.10 & 1,842.74 & 20,482.41  \\
Radar & 2.50 & 22.96 & 195.07 & 3,869.23 & 12,184.29 & 9,746.48 & OOM & OOM  \\
\hline
AEGIS & 57.43 & 1,552.91 & 3,232.93 & 4,058.93 & 7,099.13 & 6,158.25 & 13,992.45 & OOM \\
MLPAE & 50.24 & 49.48 & 65.57 & 68.66 & 95.88 & 88.73 & 148.11 & 710.24  \\
DOMINANT & 0.45 & 0.46 & 0.65 & 0.72 & OOM & OOM & OOM & OOM  \\
GAD-NR & 2,207.40 & 14,008.06 & 35,196.88 & 43,968.26 & 79,444.68 & 87,204.88 & 179,388.82 & 840,485.61 \\
\hline
CoLA & 848.14 & 1,052.65 & 2,587.15 & 2,114.03 & 4,602.92 & 4,121.26 & 10,205.07 & OOM \\
ANEMONE & 856.40 & 1,101.61 & 2,704.56 & 2,299.42 & 4,886.12 & 4,228.13 & 10,106.70 & OOM \\
SL-GAD & 1,676.18 & 2,231.07 & 5,317.08 & 4,557.71 & 9,765.55 & 8,548.84 & 20,207.68 & OOM \\
GRADATE & 895.81 & 1,408.88 & 3,375.16 & 3,010.19 & OOM & OOM & OOM & OOM \\
\hline
\rowcolor[HTML]{E9E9E9}
CMUCL & 273.04 & 873.11 & 1,752.81 & 2,395.52 & 3,791.07 & 3,158.90 & 5,770.94 & 86,443.99 \\
    \bottomrule
  \end{tabular}
\end{table*}

\end{document}